\documentclass[journal]{IEEEtran}

\usepackage{amsmath,amssymb,amsfonts}
\usepackage{algorithmic}
\usepackage{algorithm}
\makeatletter
\@ifundefined{algorithm*}{%
  \newenvironment{algorithm*}{\@dblfloat{algorithm}}{\end@dblfloat}%
}{}
\makeatother

\usepackage{graphicx}
\usepackage{bm}
\usepackage{booktabs}
\usepackage{placeins}
\usepackage{hyperref}
\usepackage{xcolor}
\usepackage[normalem]{ulem}
\usepackage[caption=false,font=footnotesize]{subfig}
\newcommand{\R}{\mathbb{R}}

\newcommand{\SEtwo}{\mathrm{SE}_2(3)}
\newcommand{\SOthree}{\mathrm{SO}(3)}
\newcommand{\SO}{\mathrm{SO}(3)}
\newcommand{\cN}{\mathcal{N}}
\newcommand{\Exp}{\mathrm{Exp}}

\newcommand{\bp}{\bm{p}}
\newcommand{\bv}{\bm{v}}
\newcommand{\bg}{\bm{g}}
\newcommand{\bn}{\bm{n}}
\newcommand{\bN}{\bm{N}}
\newcommand{\bS}{\bm{S}}
\newcommand{\bR}{\bm{R}}    
\newcommand{\bP}{\bm{P}}
\newcommand{\bH}{\bm{H}}
\newcommand{\bK}{\bm{K}}
\newcommand{\bI}{\bm{I}}
\newcommand{\bA}{\bm{A}}
\newcommand{\bL}{\bm{L}}
\newcommand{\bQ}{\bm{Q}}
\newcommand{\skewmat}[1]{[#1]_{\times}}

\begin{document}

\title{Proprioceptive Simultaneous Base and Terrain Elevation Estimation for Legged Robot Locomotion}

\author{Michael Maravgakis$^{1,\dagger}$, Stylianos Piperakis$^{2,\dagger}$ and Panos Trahanias$^{1}$%
\thanks{$^{\dagger}$These authors contributed equally to this work.}%
\thanks{$^{1}$Michael Maravgakis and Panos Trahanias are with the Institute of Computer Science, Foundation for Research and Technology - Hellas (FORTH), Heraklion, Greece.
    {\tt\small \{maravgakis,trahania\}@ics.forth.gr}}%
\thanks{$^{2}$Stylianos Piperakis is with VinDynamics.
    {\tt\small v.stylianos@vindynamics.net}}%
}

\maketitle

\begin{abstract}
Legged robots in visually degraded environments must rely exclusively on
proprioceptive sensing, under which global position and yaw are
unobservable. Vertical drift is particularly detrimental, as it corrupts
the height reference required by swing-leg trajectories and can
destabilize the robot. We treat the terrain itself as a persistent
proprioceptive landmark: every footstep provides a height measurement
that anchors the vertical axis. We propose Simultaneous Base and
Elevation Estimation (\textit{SBEE}), a real-time open-source framework
in which a Left-Invariant Extended Kalman Filter (LI-EKF) for
floating-base estimation and a grid-based elevation map are coupled in a
bidirectional loop, mirroring SLAM with terrain elevation in place of
landmarks. Stable foot contacts insert uncertainty-aware height samples
into the map, and previously mapped cells are fed back as
uncertainty-weighted vertical pseudo-measurements that act analogously to
loop closures. Unlike contact-aided invariant filters, the LI-EKF consumes
a kinematic velocity measurement instead of augmenting the state with a
contact model per leg, so its state dimension is independent of the
number of legs. We evaluate SBEE in MuJoCo on quadrupedal and bipedal
platforms and on the ANYmal-D GrandTour dataset over 296~m of
locomotion, reducing the absolute trajectory error by 22.1\% relative to
the best of three state-of-the-art proprioceptive estimators.

\end{abstract}


\section{Introduction}
\label{sec:intro}

\IEEEPARstart{L}{egged}
robots have been at the cutting-edge of robotics research for  years,
and recently, they have started proving their advanced capabilities in
industrial workplaces. This success is attributed to their  ability to
navigate and operate alongside humans in unstructured environments.
By discretizing space through consecutive steps, legged robots can
traverse obstacles, stairs, and other space discontinuities that traditional
wheeled platforms are inherently incapable of navigating.

While discrete foot-ground contacts enable this mobility,
they also pose significant challenges for autonomous operation.
Feedback controllers require continuous, high-rate information to maintain balance; robust locomotion therefore
relies strictly on accurate, high-frequency state estimation.
In visually degraded conditions or
on platforms lacking exteroceptive sensors, estimators must rely entirely
on proprioceptive sensing. As the robot walks over time, relying purely on
this relative proprioception causes the robot’s global pose to
drift arbitrarily from reality, eventually destabilizing the locomotion
controller. This drift is attributed mainly to unmodeled dynamic phenomena,
such as slippage, external forces and collision with obstacles as well as
noisy sensor measurements and additive IMU biases that build up over time.
The obvious solution is an external reference to correct such errors, but by assumption,
there is none available in the very conditions that make proprioceptive estimation
necessary.

Available information to reduce this drift is already present, but it is usually
discarded. Every footstep provides a height measurement expressed in the robot's base frame.
Once these measurements are retained, the already contacted terrain can serve as a reference point.
Similarly, when lacking external optical references, humans construct an internal
mental map of their surroundings using strictly proprioceptive and tactile
feedback. Each step yields a noisy height measurement, which is subconsciously
fused to calibrate a vertical state estimate. Upon revisiting the same location,
the familiar terrain functions as an absolute spatial anchor. Any uncertainty
in the internal sense of elevation is immediately bounded and corrected by the
mismatch between the anticipated floor height and the physical contact.

\begin{figure}
  \centering
  \includegraphics[width=\columnwidth]{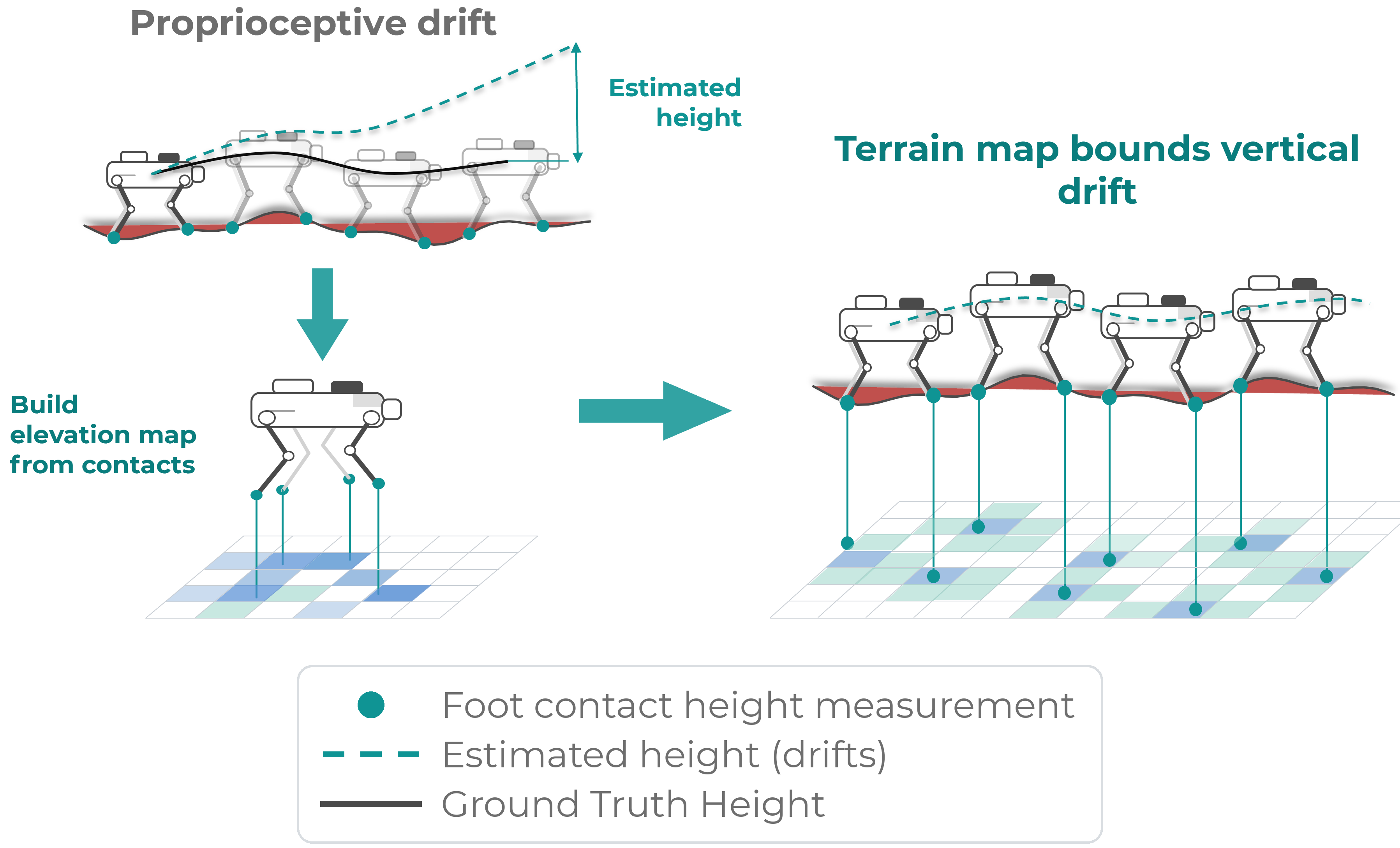}
  \caption{Proprioceptive elevation mapping using contact measurements concept illustration.}
  \label{fig:concept_illustration}
\end{figure}

This paper formalizes that intuition for legged robots
with either flat or point feet. We propose a real-time proprioceptive state estimation
framework that jointly estimates the robot's base state and a terrain elevation map using
inertial, force, and kinematic measurements. By acting as the robotic equivalent of
walking in the dark, this framework effectively anchors the unobservable vertical drift, as graphically illustrated
in Fig.~\ref{fig:concept_illustration},
facilitating real-time feedback controllers with continuous, smooth, and globally consistent
state estimates.

\section{Related Work}

\subsection{Proprioceptive and Contact-Aided Base State Estimation}

Accurate and real-time base state estimation is fundamental for legged locomotion,
as the estimated state closes the loop of the controller which acts on these high-rate estimates.
To this end, legged locomotion controllers
often rely on proprioceptive sensing, namely inertial measurements, joint encoders,
and contact information. The most widely used approach is to fuse IMU measurements
with leg kinematics utilizing nonlinear Kalman Filter (KF) variants, using stance
feet as constraints on the robot's motion.
Bloesch et al.~\cite{bloesch2012state} proposed a proprioceptive
state estimator for point-feet robots based on the fusion of leg kinematics
and IMU measurements. This formulation was later extended to address
estimation on unstable and slippery terrain by incorporating uncertainty
in the contact constraints~\cite{bloesch2013unstable}. Similar ideas were
applied to humanoid robots by Rotella et al.~\cite{rotella2014state},
who extended proprioceptive state estimation to flat-foot contacts by
including rotational constraints from the support feet. Moreover,
versatile EKF-based architectures such as PRONTO~\cite{pronto}
have generalized these proprioceptive fusion strategies to provide
reliable legged odometry across various bipedal and quadrupedal platforms.
Furthermore, cascade state estimation frameworks like SEROW~\cite{serow}
have expanded on these methods to enable estimation of 3D Center of Mass (CoM)
and base tracking on uneven terrains.
Beyond single-filter formulations, modular fusion stacks combine several estimators
into a deployable pipeline. MUSE~\cite{muse2025} cascades a nonlinear attitude observer
with an exogenous KF for orientation, a leg odometry module that derives base velocity from kinematics
and sensor-fusion KF that outputs the final pose and twist.
These methods demonstrate that contact information can effectively
constrain the floating-base state. However, in purely proprioceptive
estimation, global position and yaw remain unobservable~\cite{bloesch2012state}.
The z-axis experiences greater drift because even microscopic errors in canceling out the constant 1g pull of gravity are magnified quadratically during the double integration of acceleration.

\subsection{Invariant Filtering for Legged Robot State Estimation}

Invariant filtering has recently become a prominent approach for legged robot
state estimation because it exploits the Lie group structure of inertial
navigation and contact-aided dynamics. The theoretical basis of the invariant
EKF was established by Barrau and Bonnabel~\cite{invariantekf2017},
who proved that for suitable systems on Lie groups the invariant error
dynamics can become autonomous, in the sense that the error evolves according to a differential
equation depending only on the error itself and on the measured input, leading to improved stability compared
with standard EKF linearizations.
Hartley et al.~\cite{hartley2018contact} applied this theory to legged robots
by deriving a contact-aided invariant EKF for a Cassie bipedal robot,
where IMU propagation is corrected using forward kinematic contact measurements. Extending this line of research, subsequent work~\cite{hartley2020contact}
presented a framework for robot state estimation, including contact-inertial
dynamics, contact switching, bias estimation, and both world-centric and
robot-centric formulations. Learned contact-event estimation has also been
combined with invariant filtering to reduce reliance on dedicated contact
sensors while preserving proprioceptive operation~\cite{lin2022legged}.
Smoothing extends invariant filtering by optimizing over a window of past states
rather than only the most recent one. Yoon et al.~\cite{InvariantSmoother2024} proposed Invariant Smoother (IS),
which extended invariant filtering into a fixed-lag MAP problem, retaining state-independent
Jacobians for the propagation and kinematic-contact residuals. Marginalization keeps
the window fixed in size, so contacts older than the lag are absorbed into the prior and
can no longer be revisited.



\subsection{Elevation and Terrain Mapping}
Elevation maps are widely used for legged locomotion because they provide
a representation of terrain geometry suitable for foothold selection and
terrain-aware control. Most elevation maps are represented as 2.5-D height
maps with each horizontal location containing one elevation value $z = h(x,y)$.
Most existing elevation mapping pipelines rely on exteroceptive sensing
such as LiDAR, stereo, or depth cameras
(\cite{Fankhauser2018,miki2022elevation,belter2016rgbd,fallon2015terrain}).
Probabilistic elevation mapping methods, such as~\cite{Fankhauser2018},
have addressed the uncertainty induced by noisy range measurements and drifting
robot localization.  More specifically, they formulated robot-centric
elevation mapping for mobile robots with uncertain localization,
propagating pose uncertainty into the map representation.

Compared with exteroceptive terrain mapping, proprioceptive elevation mapping has received considerably less attention, with most existing approaches focusing on terrain classification. Yang et al.~\cite{proprio_gait_planning} proposed a proprioception-based gait-planning approach for extreme terrain traversal, where unexpected elevation changes are inferred through contact events and used to modify contact schedules and foothold plans online. This work demonstrates the value of proprioceptive information for responding to unknown terrain height variations, but the generated map is used for gait adaptation rather than continuous base state estimation.

More recently, Yang et al.~\cite{yang_elevation} proposed a terrain-aware framework that constructs a 2.5-D terrain map
from IMU, encoder, and force information, utilizing it within a Control Barrier Function (CBF) Model Predictive Controller (MPC)
to enforce global and local safety constraints.
Notably, they couple this terrain estimate with their state estimator to improve base position tracking. However, their approach fits global support-plane parameters to the map and injects these planar constraints into a standard EKF. While effective for their primary objective of safety-critical control, this planar abstraction discards fine-grained terrain geometry.

Sanchez-Delgado et al.~\cite{sanchez_elevation} also proposed a proprioceptive terrain-mapping framework for quadruped robots operating in planetary permanently shadowed regions. Their method incrementally constructs a multi-layer 2.5-D grid map containing elevation, slippage, cost of transport, and stability-related terrain metrics from internal sensing. This work further supports the idea that proprioceptive sensing can provide useful terrain information in visually degraded environments. However, its goal is terrain characterization for exploration, rather than using the elevation map as a high-rate correction source for base state estimation.

In summary, prior contact-aided estimators provide high-rate proprioceptive base-state estimates
but do not build an explicit terrain elevation memory.
Our method closes the loop that both lines of work leave open. Prior proprioceptive terrain mapping
approaches treat the map as a product of the state estimate, while the proposed approach
treats it as an active source of correction for it. We introduce Simultaneous Base and Elevation Estimation
concept coined as \emph{SBEE}, where foot contact update  a probabilistic local elevation map,
and the map in turn provides uncertainty-weighted vertical measurements to the base-state
estimator whenever the robot contacts previously observed terrain.

\subsection{Contributions}

Our estimator follows the invariant-filtering line by representing the base state on the Lie group
$\mathrm{SE}_2(3)$ with a left-invariant error formulation. Unlike classical contact-aided invariant filters,
which augment or update the state using the current contact points, our method additionally maintains a terrain
elevation map outside the filter state. Our contributions are threefold:
\begin{itemize}
    \item \emph{Simultaneous Base and Elevation Estimation (SBEE)}: to the best of our knowledge, this is the first formulation of proprioceptive terrain
        mapping as a loop-closure vertical correction for legged robot state estimation. Feeding previously
        observed terrain heights back to the base-state estimator anchors the otherwise unobservable vertical
        drift without any exteroceptive sensing, and applies to both point-foot and flat-foot morphologies.

  \item \emph{Fixed-size invariant filter without contact-state augmentation}: existing approaches augment the
        state with contact points or contact poses and model their evolution as a random
        walk~\cite{bloesch2012state, rotella2014state}, a poor approximation for rigid ground
        contacts~\cite{yang2025multi} that introduces additional parameters to tune. Using the map as an external, uncertainty-aware memory keeps the state dimension independent of the number of active contacts, reducing both computational and tuning cost, and allowing the formulation to apply unchanged to platforms with an arbitrary number of legs. The map is fused through a Schmidt--Kalman partial update~\cite{brink2017partial} that corrects the vertical base position only, so an external and possibly erroneous terrain memory can never corrupt the observable velocity, attitude and bias states.

  \item \emph{Real-time proprioceptive mapping}: a probabilistic grid-based elevation map with toroidal indexing
        for $O(1)$ index translation, allowing the map to recenter without shifting accumulated data in memory,
        combined with inverse-distance-weighting interpolation and neighborhood propagation for rapid
        convergence.
\end{itemize}
Our framework will be released as open-source upon acceptance, built on an optimized C++17 core with native
Python bindings and ROS/ROS2 compatibility.

The remainder of this paper is organized as follows:
Section~\ref{sec:problem_formulation} formalizes the problem and establishes the notation.
Section~\ref{sec:terrain_map} presents the proprioceptive terrain elevation algorithm, while
Section~\ref{sec:liekf} details our formulation of the Left-Invariant EKF for floating-base estimation.
Section~\ref{sec:coupling} then couples the two components into the complete SBEE framework.
Section~\ref{sec:results} reports the experimental evaluation in simulation and on the ANYmal-D GrandTour
dataset, and Section~\ref{sec:discussion} examines the properties of the proposed formulation together with its
limitations. Finally, Section~\ref{sec:conclusions} concludes the paper.
\begin{figure}[!b]
  \centering
  \includegraphics[width=0.8\columnwidth]{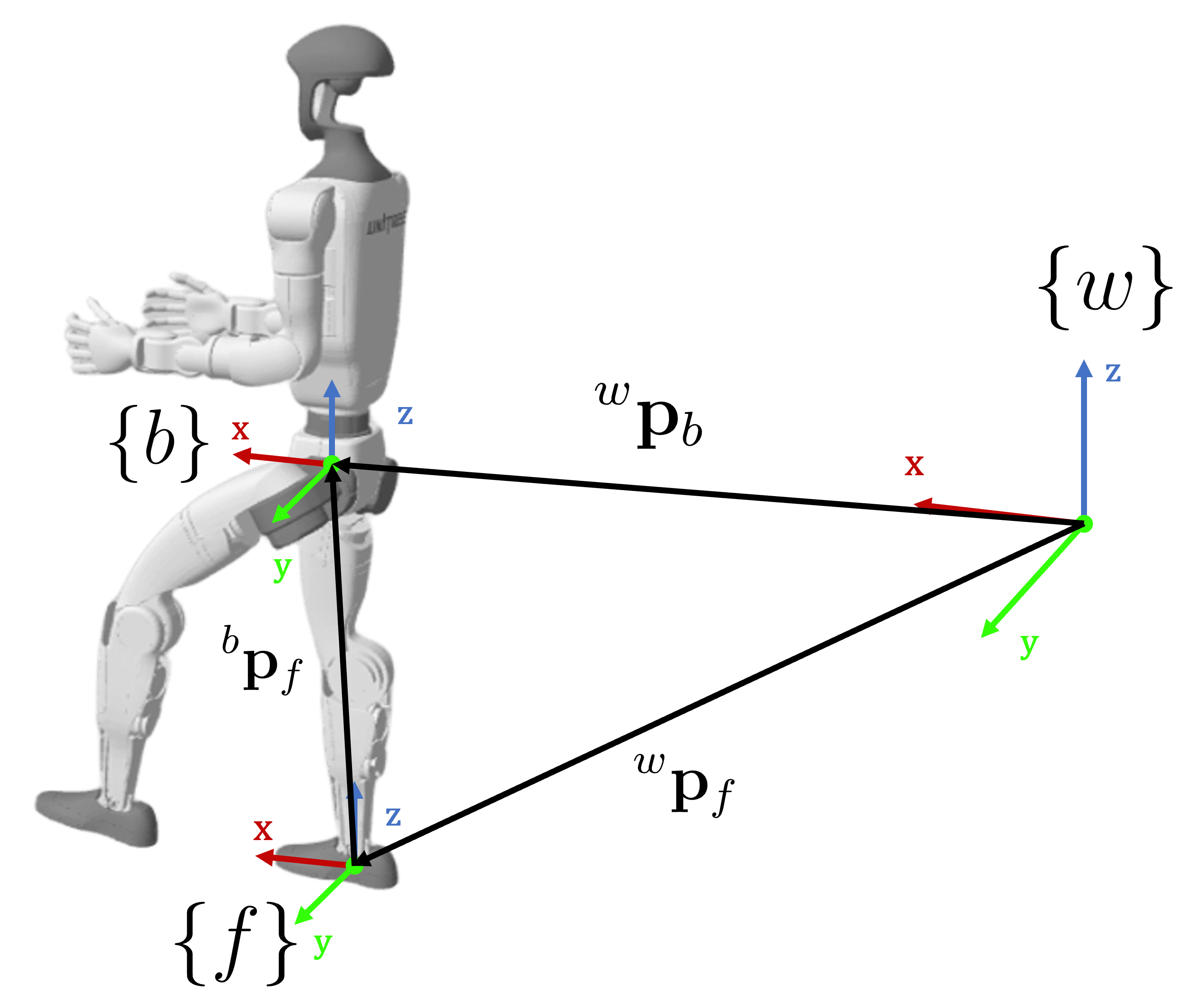}
  \caption{Coordinate frames and notation for a G1 bipedal robot.}
  \label{fig:notation}
\end{figure}


\section{Problem Formulation}
\label{sec:problem_formulation}
\subsection{Notation and Coordinate Frames}
\label{ssec:notation}
To ensure clarity throughout the remainder of this manuscript, we first define the mathematical
notation and coordinate frames used in this work. The primary coordinate systems are: $\{w\}$
the fixed inertial world frame, the moving robot base $\{b\}$ and the foot contact frame $\{f\}$
which are illustrated in Fig.~\ref{fig:notation}. Regarding variable types, all vectors are denoted by bold lowercase letters
(e.g. $\bv$, $\bp$) while scalars are denoted by non-bolded lowercase letters (e.g. $h$,  $k$).
Moreover,  matrices are denoted by bold uppercase letters (e.g. $\bR$, $\bm{\Sigma})$.
A position vector ${}^{a}\bp_{d}$ denotes the
position of point $d$ relative to the origin of frame $\{a\}$, expressed
in its coordinates, e.g.
${}^{w}\bp_b = ({}^{w}p^{x}_b,\,{}^{w}p^{y}_b,\,{}^{w}p^{z}_b)\in\R^3$
is the base position in the world frame and ${}^{b}\bp_f$ the position
of foot $f$ in the base frame. A rotation matrix
${}^{a}\bR_{c}\in SO(3)$, the three-dimensional Special Orthogonal group,
maps vectors from $\{c\}$ to $\{a\}$, e.g. ${}^{w}\bR_b$ maps vectors from
the base frame into the world frame. Frame indices are omitted where
unambiguous, in which case world frame quantities are implied.

Furthermore,  the hat operator $(\cdot)^{\wedge}$ is used to map the higher-dimensional state perturbation to its corresponding Lie Algebra  element in $\mathfrak{se}_2(3)$. In contrast, the skew-symmetric matrix operator $\skewmat{\cdot}$ is strictly reserved for mapping vectors in $\mathbb{R}^3$ to $\mathfrak{so}(3)$ allowing the standard cross product to be computed via matrix multiplication as  $\bm{a} \times \bm{b} = \skewmat{\bm{a}} \bm{b}$. The notations $\bm{0}_n$ and $\bm{I}_n$ with $n \in \mathbb{N}$ denote the $n$-dimensional zero and identity matrix respectively, while $\bm{0}_{n \times m}$ denotes an $n \times m$ zero matrix.

Finally, regarding the covariance matrices, four distinct symbols are utilized to represent the level of uncertainty. Lowercase $\sigma^2$ denotes scalar variances that are used exclusively in the terrain elevation mapping.  $\bm{\Sigma}$ denotes  sensor-level noise covariances associated with physical measurement devices. $\bQ$ is reserved for filter process noise covariance and captures the uncertainty injected into the full filter state. Finally, $\bm{N}$ denotes the effective noise covariance at the update step of the filter.

\subsection{The Unobservability of Vertical Base Position}
\label{ssec:z_unobs}
In legged robot state estimation relying solely on proprioceptive sensing (IMU, force sensors and joint encoders),
the global translation (${}^{w}p^{x}_b, {}^{w}p^{y}_b, {}^{w}p^{z}_b)$ and the yaw angle are fundamentally unobservable.
However, the consequences of this property differ drastically across dimensions. Drifting in the horizontal plane is
generally benign and acts as a shift in the local reference frame which modern locomotion controllers can accommodate.
On the contrary, drifting in the vertical position $({}^{w}p^{z}_b)$ is physically catastrophic because legged locomotion
relies strictly on predictable foot-ground interactions. Recovering position from an IMU sensor requires double-integration,
and any bias or noise in the measurement causes the position estimate to drift quadratically with time. Leg kinematics constrain
the relative displacement between the base and a foot in contact, but without an absolute altitude reference
the global ${}^{w}p^{z}_b$ remains unanchored. Over a long walk, the estimated height of the robot can diverge from reality.
An upward drift of ${}^{w}p^{z}_b$ makes the terrain appear farther below the base than it actually is, causing the swing-legs
to overextend and to produce a large unmodeled impact with the ground.
A downward drift has the opposite effect, causing the touchdown to be delayed or missed.
In both cases the timing and the location of the contact event are corrupted which can potentially lead to loss of balance and eventually a fall.




\subsection{Overview}
\label{ssec:overview}

\begin{figure*}[!t]
  \centering
  \includegraphics[width=\textwidth]{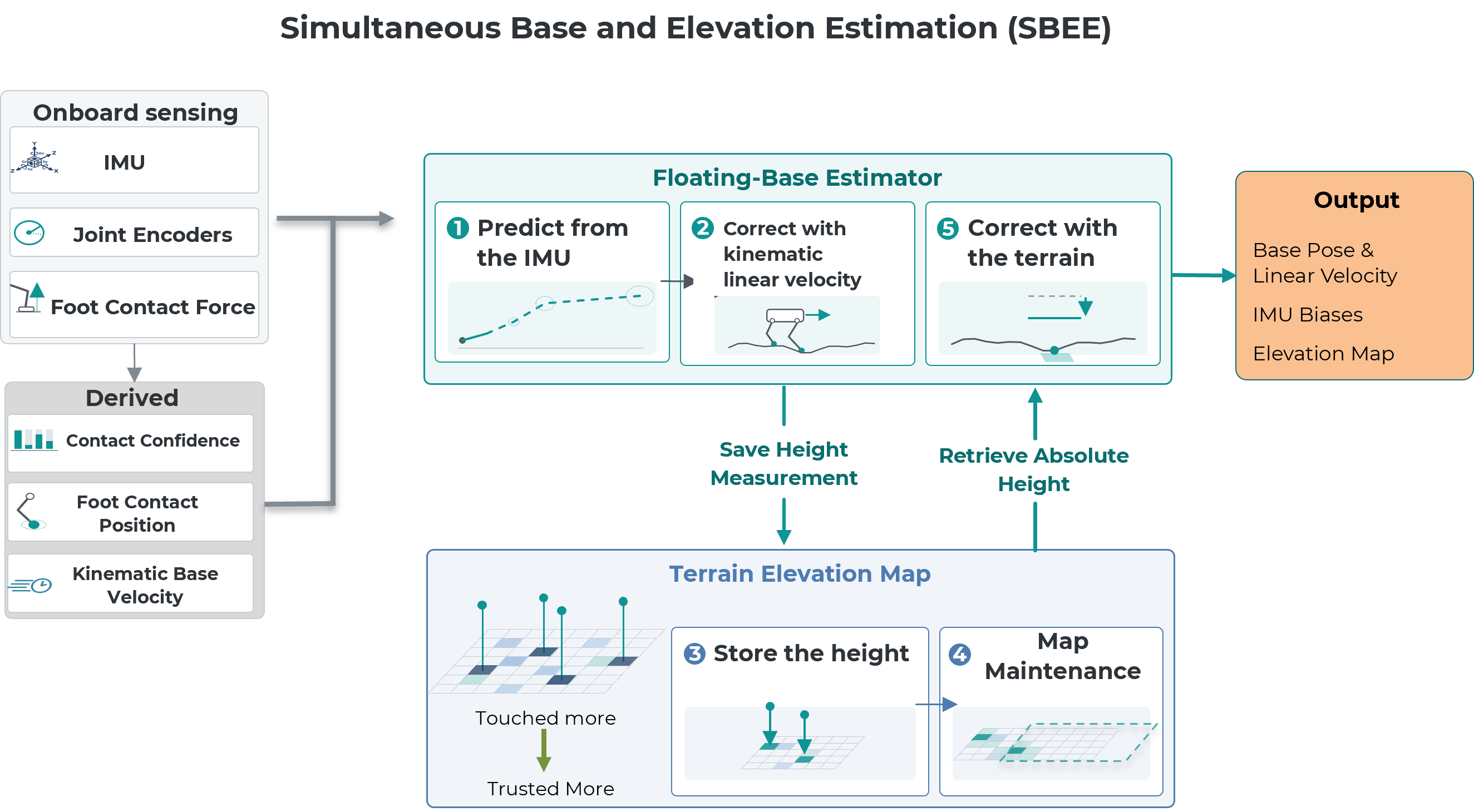}
  \caption{Overview of the Simultaneous Base and Elevation Estimation
    framework. IMU and kinematic measurements update the LI-EKF, stable
    contacts update the local elevation map, and map elevations are fed back
    as uncertainty-weighted vertical world pseudo-position measurements.}
  \label{fig:flowchart}
\end{figure*}

Fig.~\ref{fig:flowchart} summarizes the proposed framework, which
consists of two tightly coupled components:

\begin{enumerate}
  \item \textbf{Terrain Elevation Mapper}
        (Section~\ref{sec:terrain_map}) that maintains a probabilistic
        $M \times M$ grid of terrain heights, built and refined
        incrementally from foot-contact observations.
  \item \textbf{Left-Invariant Extended Kalman Filter} (LI-EKF,
        Section~\ref{sec:liekf}) whose state lives on the Lie group
        $\SEtwo \times \R^{6}$, exploiting the group-affine structure of
        IMU-driven dynamics to obtain a state-independent linearized
        error propagation.
\end{enumerate}
The two components interact in a bidirectional loop, numbered
Step 1--Step 5 in Fig.~\ref{fig:flowchart}. At each cycle the
LI-EKF is propagated with the IMU (Step 1) and corrected with the
kinematic base velocity derived from the legs (Step 2). The
predicted base state then uses the contact information to update the map with
height measurements (Step 3), which is recentered and interpolated
as the robot moves (Step 4). Finally, the accumulated terrain
knowledge is fed back as a vertical pseudo-measurement that corrects the
filter's vertical drift (Step 5). When the robot transitions from
unmapped terrain onto previously-mapped terrain, the low-variance cells
produce corrections of the same qualitative effect as a SLAM loop
closure.

\section{Terrain Elevation Map}
\label{sec:terrain_map}
\subsection{Contact State Estimation}
\label{sec:contact_est}
Determining which feet are supporting  the robot's weight in real-time is essential for accurate state estimation. Common approaches often rely on ground reaction force (GRF) thresholds or hysteresis thresholding (e.g. Schmitt trigger) due to implementation simplicity, low computational complexity and adequate performance.
However, these techniques are unstable subject to varying payloads or unmodeled dynamic events such as  impacts and foot slippage, and require manual tuning of hyperparameters that commonly depend on the walking gait. To address this, we utilize a continuous contact confidence rather than a binary switch.
Similarly to \cite{masuya2015} the confidence of contact for each leg is estimated by normalizing its vertical GRF against the instantaneous average load across all end-effectors:
\begin{equation}
  CC_f =
  \begin{cases}
    \min\!\left(1,\ \dfrac{F_f}{\frac{1}{N}\sum_{j=1}^{N} F_j}\right), & \sum_{j=1}^{N} F_j > 0 \\[2ex]
    0,                                                                 & \text{otherwise}
  \end{cases}
  \label{eq:contact_prob}
\end{equation}
where $CC_i$ and $F_i$ are the contact confidence and measured vertical GRF of leg $i$ respectively, clamped in $[0, m g]$. This load-normalized formulation provides a dynamic certainty score bounded between 0 and 1, ensuring that the state estimator adapts smoothly during dynamic maneuvers and variable payload distributions.
Nevertheless, this simple, efficient formulation relies solely on vertical GRF, which cannot reliably capture slippage.
With additional foot-mounted sensors (e.g., IMUs), more sophisticated contact-confidence or contact-probability estimators are available~\cite{PiperakisContact, RotellaContact, MaravgakisContact}.
The proposed framework is modular with respect to the contact confidence input, meaning that any external
estimator providing a continuous contact confidence or contact probability can be utilized in place of~\eqref{eq:contact_prob} without modifying the estimator.

\subsection{Proprioceptive Terrain Sensing}
\label{ssec:tactile}

At each control cycle, we estimate a $1D$ contact confidence
$CC_f$ for every foot $f$.  Subsequently, its world frame position is given by:
\begin{equation}
  \label{eq:foot_world}
  ^{w}\bp_f = {}^{w}\bR_{b}\, {}^{b}\bp_f + {}^{w}\bp_{b}
\end{equation}
where ${}^{b}\bp_f$ is the foot contact position relative to the base frame.
For point feet, the contact point coincides with the leg end-effector.
For flat feet, we need to determine the anchoring pivot, namely the point
within the contact patch that instantaneously anchors the foot to the ground \cite{masuya2015}.

This contact point provides a height sample $z = {}^{w}\bp_{f}^{z}$. The measurement variance supplied to the map is the world frame
kinematic $z$-variance divided by the contact confidence:
\begin{equation}
  \sigma^{2}_{z}
  = \frac{\bm{e}_{3}^{\top}\,^{w}\bR_{b}\,^{b}\bm{\Sigma}_{f}\,^{w}\bR_{b}^{\top}\,\bm{e}_{3}
    + \varepsilon}
  {CC_f}
  \label{eq:meas_var}
\end{equation}
where $\bm{e}_{3} = (0,0,1)^{\top}$ is the unit vector along the world $z$-axis and $\varepsilon \ll 1$ to ensure that $\sigma^2_z > 0$ since ${}^{b}\bm{\Sigma}_{f}$ is positive semi-definite and its $z$-component vanishes near kinematic singularities such as a fully extended leg. More specifically, ${}^{b}\bm{\Sigma}_{f}$  is the Cartesian covariance matrix of the forward kinematics for foot $f$ expressed in the base frame $b$ , obtained by propagating the joint-space uncertainty into the Cartesian space via:

\begin{equation}
  {}^{b}\bm{\Sigma}_{f} = {}^{b}\bm{J}_{f}\,\bm{\Sigma}_{joints}\, {}^{b}\bm{J}^{\top}_{f}
  \label{eq:kin_cov}
\end{equation}
where ${}^{b}\bm{J}_{f}$ is the corresponding analytical leg end-effector Jacobian and $\bm{\Sigma}_{joints}$ is the diagonal covariance matrix containing the measurement noise of all the joint encoders.

\subsection{Map Representation and Toroidal Indexing}
\label{ssec:map_repr}
\label{ssec:toroidal}

The local terrain is represented as a two-dimensional grid of $M\times M$ cells centered on the robot's current
position. Each cell $c$ inside the grid stores an \emph{elevation state}
\begin{equation}
  c = \bigl(h,\;\sigma^{2}\bigr)
  \label{eq:cell}
\end{equation}
where $h \in \R$ is the estimated terrain height and $\sigma^{2} \in \R_{>0}$ is the associated variance, computed
via Kalman filtering as explained in Section~\ref{ssec:cell_update}.

Given a spatial resolution $\delta > 0$ (meters per cell), the map covers a square region of side length
$L = M\,\delta$. A location $\bm{\ell} = (\ell^x, \ell^y)^{\top}$ is discretized into a \emph{global index} $i_{g}$:
\begin{equation}
  \bm{i}_{g}(\bm{\ell})
  = \Bigl(\bigl\lfloor \delta^{-1}\,\ell^x + 0.5 \bigr\rfloor,\;
  \bigl\lfloor \delta^{-1}\,\ell^y + 0.5 \bigr\rfloor\Bigr)^{\!\top}
  \label{eq:loc2gidx}
\end{equation}
with sign-dependent rounding for a discretization that is symmetric about the origin.

To avoid expensive memory copies whenever the map is recentered, the grid is stored as a circular buffer and
addressed through \emph{toroidal} indexing, inspired by~\cite{rogmap2024}. The mapping from a global index
$\bm{i}_{g}$ to the one-dimensional storage \emph{hash id} $\eta$ proceeds through an intermediate \emph{local
  index} $\bm{i}_{l}$:
\begin{align}
  i_{l}^{k} & = \bigl(i_{g}^{k} \bmod M\bigr) \in
  \bigl[-\tfrac{M}{2},\; \tfrac{M}{2}-1\bigr],
  \quad k \in \{0,1\}
  \label{eq:g2l}                                        \\[4pt]
  \eta(\bm{i}_{l})
            & = \bigl(i_{l}^{0} + \tfrac{M}{2}\bigr)\,M
  + \bigl(i_{l}^{1} + \tfrac{M}{2}\bigr)
  \label{eq:hashid}
\end{align}
We specifically restrict $M$ to be a power of two, so that the modulo in~\eqref{eq:g2l} reduces to a bitwise AND
and the index conversion is $O(1)$. As the robot moves, the map origin is shifted by updating the
global-to-local offset: only the cells that fall outside the new boundary are reset, while all remaining
cells retain their accumulated elevation estimates without any data movement.

\subsection{Boundary Checking and Map Re-centering}
\label{ssec:recenter}

A location $\bm{\ell}$ is \emph{inside} the map if its global index
satisfies
\begin{equation}
  \bigl|i_{g}^{k} - o^{k}\bigr| \le \frac{M}{2},
  \quad k \in \{0,1\}
  \label{eq:inside}
\end{equation}
where $\bm{o}$ is the current map origin.
When the robot's base position drifts beyond $d_{\max}$ from the map origin in the horizontal plane, the map is re-centered.  Let
$\Delta\bm{o} = \bm{o}' - \bm{o}$ denote the shift.  If
$|\Delta o^k| \ge M$ for any $k$, the entire map is reset; otherwise,
the $|\Delta o^k|$ cell rows or columns that now lie outside the boundary
are cleared.  All other cells, including terrain the robot may
return to, are preserved, making it possible for the robot to
\emph{re-encounter} its own tactile history.
With this mechanism, in order to shift the entire map, we avoid copying all the surveyed cells to a new memory location.
Instead, we shift the map origin and clear the cells that now lie outside the boundary.
This is crucial for real-time and high-frequency execution.

\subsection{Cell Update via Kalman Filtering}
\label{ssec:cell_update}

Each cell maintains a scalar Gaussian belief
$c \sim \cN\!\bigl(h,\, \sigma^{2}\bigr)$ inspired from~\cite{Fankhauser2018}.
Nevertheless, instead of using exteroception to obtain a terrain elevation measurement, we use the proprioceptive measurement from the foot contact.
When a foot makes contact with the terrain, it produces a height measurement $z$ with variance
$\sigma^{2}_{z}$, the cell is updated as:
\begin{align}
  K & = \frac{\sigma^{2}}
  {\sigma^{2} + \sigma^{2}_{z}}
  \label{eq:kalman_gain}         \\[3pt]
  h^{+}
    & = h + K\,\bigl(z - h\bigr)
  \label{eq:height_update}       \\[3pt]
  \sigma^{2+}
    & = (1 - K)\,\sigma^{2}
  \label{eq:var_update}
\end{align}
with $K$ denoting the Kalman gain. Every successive contact on the same cell \emph{monotonically reduces}
the variance $\sigma^{2}$.  A cell that has been touched many
times, for example, a spot the robot has stepped on repeatedly, has
very low variance and therefore acts as a strong anchor.  This is
precisely the mechanism that enables the loop-closure effect: when the
robot returns to a well-converged cell after a period of walking on
unmapped terrain, the cell's low $\sigma^{2}$ produces a large Kalman
gain when used in the terrain correction of the SBEE framework, presented in Section~\ref{ssec:ekf_terrain}, pulling the drifted
${}^{w}p^{z}_{b}$ back to the correct value. To address the issue of filter overconfidence due to the accumulated variance ($\sigma^2$) reduction, we introduced a minimum variance value for each cell, thereby balancing map confidence to newly acquired information.

\subsection{Neighborhood Propagation}
\label{ssec:neighbourhood}

To accelerate map convergence in cells that have not been directly
touched, the update is propagated to a circular neighborhood of radius
$r$ cells. For a neighbor at displacement $(\Delta x, \Delta y)$:
\begin{align}
  z_{\text{nb}}          & = h^{+}
  - \frac{n_x}{n_z}\,\Delta x
  - \frac{n_y}{n_z}\,\Delta y
  \label{eq:plane_proj}                       \\[4pt]
  \sigma^{2}_{\text{nb}} & = \sigma^{2}_{z}\,
  \bigl(1 + \alpha\,(\Delta x^{2} + \Delta y^{2})\bigr)
  \label{eq:dist_var}
\end{align}
where $h^{+}$ is the posterior height of the contact cell,
$\bn = (n^x, n^y, n^z)^{\top}$ is the estimated surface normal
(defaulting to $n = (0 ,0, 1)^T$ for point feet robots), and $\alpha > 0$ is a
distance-dependent variance inflation gain.
Equation~\eqref{eq:plane_proj} projects the contact-cell height onto
the local tangent plane, providing first-order slope information when
the foot has a planar contact patch in \emph{stable} contact,
namely high contact confidence and low foot linear and angular velocities.

Propagation extends the robot's ``sense of touch'' beyond the
immediate footprint, so that when the robot next steps near a
previously-touched region, even cells that were not directly contacted
can provide informative corrections.

\subsection{Inverse Distance Weighting Interpolation}
\label{ssec:idw}

Between update cycles, \emph{Inverse Distance Weighting} (IDW)
interpolation fills cells within the bounding box of recent $N_c$ contact
points:
\begin{equation}
  \hat{h}(\bm{\ell})
  = \frac{\displaystyle\sum_{j=1}^{N_c} w_j\, h^{(j)}}
  {\displaystyle\sum_{j=1}^{N_c} w_j},
  \qquad
  w_j = \frac{1}{\|\bm{\ell} - \bm{\ell}^{(j)}\|^{p}}
  \label{eq:idw}
\end{equation}
with  $p$ the power hyperparameter.  Variance is interpolated similarly.
This extends the mapped region between the feet, filling map gaps and increasing the
likelihood that a future contact will land on a cell that already
carries information.


\section{Left-Invariant Extended Kalman Filter}
\label{sec:liekf}
For three-dimensional floating-base estimation, we employ a
Left-Invariant Extended Kalman Filter (LI-EKF)~\cite{invariantekf2017}.
In contrast to a conventional EKF, which defines the estimation error
through direct vector subtraction, the invariant EKF defines the estimation
error multiplicatively on the underlying matrix Lie group. The resulting
error is invariant and is represented in the
local robot frame. 

The choice between the left- and right-invariant error conventions is
dictated by the form of the available measurement. The proposed filter
consumes a kinematic base-velocity measurement expressed in the world
frame (Section~\ref{ssec:kin_update}), and world frame quantities admit the
left-invariant observation form of~\cite{invariantekf2017}. Pairing such an
observation with the left-invariant error yields a state-independent
observation matrix, which the right-invariant convention would not provide
for this measurement. The terrain elevation measurement is a scalar world frame height and is incorporated as a partial scalar update in
the same error coordinates (Section~\ref{ssec:ekf_terrain}).

For the bias-free system, the process dynamics are
group-affine. Consequently, the left-invariant estimation error evolves
autonomously, independently of the true and estimated base trajectories,
and its logarithm satisfies a linear differential equation. This is the
\emph{log-linear} property established for invariant filtering
in~\cite{invariantekf2017} and applied to legged-robot state estimation
in~\cite{hartley2020contact}. However, the gyroscope and
accelerometer biases are crucial in estimation,
hence the proposed estimator appends them as Euclidean states in Section~\ref{ssec:imubiases}.
This conventional bias augmentation breaks the exact group-affine structure of the complete
system, and the resulting estimator is therefore an ``imperfect'' invariant EKF.
Nevertheless, in our left invariant formulation, the state propagation Jacobian remains
independent of the estimated orientation, velocity, and position and
depends only on the bias-corrected IMU measurements.

Unlike contact-aided invariant filters that augment the state with a
position for every active contact~\cite{hartley2020contact}, the proposed
filter uses a three-dimensional base-velocity measurement obtained from
leg kinematics, following the general idea of~\cite{kuindersma2016}. The
filter state therefore has a fixed dimension, independently of the number
of active contacts and their geometry, namely point or flat feet contacts.
This reduces computational cost and the number of parameters
that require manual tuning.

During ground contact, a world frame kinematic measurement
${}^{w}\bv_b^{\mathrm{kin}}$ of the base linear velocity is constructed
using an orientation estimate ${}^{w}\bR_b^{\mathrm{imu}}$ obtained from
the IMU attitude estimator~\cite{mahony2008}, together with the kinematic
relation between the base and each active foot--ground contact.

For each active contact $f$, the base-frame velocity estimate is:
\begin{equation}
  {}^{b}\bv_{b,f}^{\mathrm{kin}}
  =
  -{}^{b}\dot{\bp}_{f}
  -
  \bm{\omega}^{\mathrm{imu}}
  \times {}^{b}\bp_{f}
  \label{eq:single_contact_velocity}
\end{equation}
Here, $\bm{\omega}^{\mathrm{imu}}\times{}^{b}\bp_f$ is the
tangential foot velocity induced by base rotation.
The individual estimates are combined using the normalized contact
confidence:
\begin{equation}
  W_{CC_f}
  =
  \frac{CC_f}{\sum_k CC_k},
  \qquad
  \sum_f W_{CC_f}=1
  \label{eq:contact_probability_weight}
\end{equation}
resulting in:
\begin{align}
  {}^{w}\bv_b^{\mathrm{kin}}
   & =
  {}^{w}\bR_b^{\mathrm{imu}}
  \sum_f W_{CC_f}
  \left(
  -{}^{b}\dot{\bp}_{f}
  -
  \bm{\omega}^{\mathrm{imu}}
  \times{}^{b}\bp_f
  \right)
  \label{eq:kin_vel_measurement}
  \\[2mm]
  {}^{w}\bm{\Sigma}_{v}^{\mathrm{kin}}
   & =
  \sum_f W_{CC_f}
  \Bigl(
  {}^{w}\bR_b^{\mathrm{imu}}\,
  {}^{b}\bm{\Sigma}_{f}\,
  {}^{w}\bR_b^{\mathrm{imu}\top}
  \nonumber        \\
   & \hspace{20mm}
  +
  \skewmat{
    {}^{w}\bR_b^{\mathrm{imu}}\,
    {}^{b}\bp_f
  }
  {}^{w}\bQ_{\bm{\omega}}
  \skewmat{
    {}^{w}\bR_b^{\mathrm{imu}}\,
    {}^{b}\bp_f
  }^{\top}
  \Bigr)
  \label{eq:kin_vel_covariance}
\end{align}
where ${}^{b}\bm{\Sigma}_{f}$ is the kinematic linear-velocity uncertainty of
contact $f$, and ${}^{w}\bQ_{\bm{\omega}}$ is the gyroscope noise
spectral density expressed in the world frame. The resulting
${}^{w}\bm{\Sigma}_{v}^{\mathrm{kin}}$ is treated as a continuous-time
spectral density and is converted into a discrete measurement covariance
at the LI-EKF update step. Notice, that we don't need extra tuning parameters for the kinematic velocity
covariance as it is directly derived from the contact confidence, the foot kinematics and the IMU/joint encoders noise.

\subsection{State Space and Left-Invariant Error}
\label{ssec:state_space}

The robot's base state is represented on the extended special Euclidean
group $\SEtwo$:
\begin{equation}
  \mathcal{X}
  =
  \begin{pmatrix}
    \bR               & \bv & \bp \\
    \bm{0}_{1\times3} & 1   & 0   \\
    \bm{0}_{1\times3} & 0   & 1
  \end{pmatrix}
  \in\SEtwo
  \label{eq:group_state}
\end{equation}
where $\bR\in\SO$ rotates vectors from the base frame to the world frame,
while $\bv,\bp\in\R^3$ denote the world frame base linear velocity and
position, respectively, where we dropped the frame indices for conciseness.
Let $\mathcal{X}$ denote the true state and
$\hat{\mathcal{X}}$ its estimate. The left-invariant group error is
defined as:
\begin{equation}
  \bm{\eta}^{L}
  =
  \mathcal{X}^{-1}\hat{\mathcal{X}}
  \label{eq:left_invariant_error}
\end{equation}
Expanding~\eqref{eq:left_invariant_error} gives:
\begin{equation}
  \bm{\eta}^{L}
  =
  \begin{pmatrix}
    \bR^{\top}\hat{\bR}
                      &
    \bR^{\top}(\hat{\bv}-\bv)
                      &
    \bR^{\top}(\hat{\bp}-\bp)
    \\
    \bm{0}_{1\times3} & 1 & 0 \\
    \bm{0}_{1\times3} & 0 & 1
  \end{pmatrix}
  \label{eq:left_error_expanded}
\end{equation}
Thus, the left-invariant orientation, velocity, and position errors are
represented in the local base coordinates.

The group error is parameterized by the Lie-algebra perturbation
$\bm{\xi} \in \mathbb{R}^9$:
\begin{equation}
  \bm{\eta}^L = \mathrm{Exp}\left(\bm{\xi}\right) \quad \text{and} \quad
  \bm{\xi} = \begin{pmatrix} \bm{\xi}_R \\ \bm{\xi}_v \\ \bm{\xi}_p \end{pmatrix}
\end{equation}
where $\mathrm{Exp}(\cdot)$ denotes the exponential map of $SE_2(3)$,
defined as $\mathrm{Exp}(\bm{\xi}) \triangleq \mathrm{exp_m}(\bm{\xi}^\wedge)$
with $\mathrm{exp_m}(\cdot)$ the corresponding matrix exponential, and the hat
operator maps $\bm{\xi}$ to its corresponding Lie-algebra element:
\begin{equation}
  \bm{\xi}^\wedge = \begin{pmatrix}
    [\bm{\xi}_R]_\times & \bm{\xi}_v & \bm{\xi}_p \\
    \bm{0}_{1\times3}   & 0          & 0          \\
    \bm{0}_{1\times3}   & 0          & 0
  \end{pmatrix} \in \mathfrak{se}_2(3)
\end{equation}
For the rest of this manuscript, the same symbol $\mathrm{Exp}(\cdot)$ is used for the exponential maps of
both $SO(3)$ and $SE_2(3)$; the intended group follows from the dimension
of its argument.

\subsection{Group-Affine Dynamics and Autonomous Error Evolution}
\label{ssec:group_affine_proof}
The IMU measurements are modeled as additive Gaussian processes:
\begin{align}
  \bm{\omega}_{t}^{\mathrm{imu}}
   & =
  \bm{\omega}_{t}
  +
  \bm{w}_t^{g},
  \qquad
  \bm{w}_t^{g}\sim\cN\!\bigl(\bm{0},\,\bQ_{\omega}\bigr)
  \label{eq:imu_gyro_noise}
  \\
  \bm{a}_{t}^{\mathrm{imu}}
   & =
  \bm{a}_{t}
  +
  \bm{w}_t^{a},
  \qquad
  \bm{w}_t^{a}\sim\cN\!\bigl(\bm{0},\,\bQ_{a}\bigr)
  \label{eq:imu_acc_noise}
\end{align}
and the system input is:
\begin{equation}
  \bm{u}_{t}
  =
  \begin{pmatrix}
    \bm{\omega}_{t} \\
    \bm{a}_{t}
  \end{pmatrix}
  \label{eq:imu_input}
\end{equation}
The continuous-time dynamics are:
\begin{align}
  \frac{d}{dt}\bR_{t}
   & =
  \bR_{t}\,
  \skewmat{
  \bm{\omega}_{t}^{\mathrm{imu}}
  -
  \bm{w}_t^{g}
  }
  \label{eq:continuous_R}
  \\
  \frac{d}{dt}\bv_{t}
   & =
  \bR_{t}
  \bigl(
  \bm{a}_{t}^{\mathrm{imu}}
  -
  \bm{w}_t^{a}
  \bigr)
  +
  \bg
  \label{eq:continuous_v}
  \\
  \frac{d}{dt}\bp_{t}
   & =
  \bv_{t}
  \label{eq:continuous_p}
\end{align}
These can be rewritten compactly on $\SEtwo$ as:
\begin{align*}
  \frac{d}{dt}\mathcal{X}_{t}
   & =
  \begin{pmatrix}
    \bR_{t}\,\skewmat{\bm{\omega}_{t}^{\mathrm{imu}}} & \bR_{t}\,\bm{a}_{t}^{\mathrm{imu}}+\bg & \bv_{t} \\
    \bm{0}_{1\times3}                                 & 0                                      & 0       \\
    \bm{0}_{1\times3}                                 & 0                                      & 0
  \end{pmatrix} \\
   & \quad -
  \mathcal{X}_{t}
  \begin{pmatrix}
    \skewmat{\bm{w}_t^{g}} & \bm{w}_t^{a} & \bm{0}_{3 \times 1} \\
    \bm{0}_{1\times3}      & 0            & 0                   \\
    \bm{0}_{1\times3}      & 0            & 0
  \end{pmatrix}
\end{align*}
or equivalently:
\begin{equation}
  \frac{d}{dt}\mathcal{X}_{t}
  =
  f_{\bm{u}_{t}}(\mathcal{X}_{t})
  -
  \mathcal{X}_{t}\,
  \bm{w}_t^{\wedge},
  \quad
  \bm{w}_t
  =
  \begin{pmatrix}
    \bm{w}_t^{g} &
    \bm{w}_t^{a} &
    \bm{0}_{3 \times 1}
  \end{pmatrix}^\top
  \label{eq:group_dynamics_compact}
\end{equation}
where $f_{\bm{u}_{t}}(\cdot)$ collects the deterministic group dynamics driven by the IMU input. Equation~\eqref{eq:group_dynamics_compact} satisfies the group-affine
property as proved in~\cite{invariantekf2017}. Consequently, the left-invariant error is state-trajectory independent
during propagation and obeys:
\begin{align}
  \frac{d}{dt}\bm{\eta}_{t}^{L}
   & =
  g_{\bm{u}_{t}}^{L}(\bm{\eta}_{t}^{L})
  +
  \bm{w}_{t}^{\wedge}\,
  \bm{\eta}_{t}^{L}
  \label{eq:autonomous_left_error}
\end{align}
where
\begin{align}
  g_{\bm{u}_{t}}^{L}(\bm{\eta}_{t}^{L})
   & =
  f_{\bm{u}_{t}}(\bm{\eta}_{t}^{L})
  -
  f_{\bm{u}_{t}}(\bm{I})\,
  \bm{\eta}_{t}^{L}
  \label{eq:left_error_function}
\end{align}
From the log-linear property~\cite{invariantekf2017}, the left-invariant
error $\bm{\eta}_{t}^{L}=\Exp(\bm{\xi}_{t})$ satisfies:
\begin{equation}
  \frac{d}{dt}\bm{\xi}_{t}
  =
  \bA_{t}\,\bm{\xi}_{t}
  +
  \bm{w}_{t}
  \label{eq:log_linear_property}
\end{equation}
with $\bA_{t}$ being the $9\times9$ state Jacobian:
\begin{equation}
  \bA_{t}
  =
  \begin{pmatrix}
    -\skewmat{\bm{\omega}_{t}^{\mathrm{imu}}}
     & \bm{0}_{3}
     & \bm{0}_{3}
    \\
    -\skewmat{\bm{a}_{t}^{\mathrm{imu}}}
     & -\skewmat{\bm{\omega}_{t}^{\mathrm{imu}}}
     & \bm{0}_{3}
    \\
    \bm{0}_{3}
     & \bI_{3}
     & -\skewmat{\bm{\omega}_{t}^{\mathrm{imu}}}
  \end{pmatrix}
  \label{eq:AX_left}
\end{equation}
Since
$\bm{\eta}_{t}^{L}=\Exp(\bm{\xi}_{t})
  \approx
  \bm{I}+\bm{\xi}_{t}^{\wedge}
  +\mathcal{O}(\lVert\bm{\xi}_{t}\rVert^{2})$, then:
\begin{equation}
  \frac{d}{dt}\bm{\eta}_{t}^{L}
  =
  \frac{d}{dt}\bm{\xi}_{t}^{\wedge}
  \label{eq:first_order_error_derivative}
\end{equation}
Substituting~\eqref{eq:first_order_error_derivative}
into~\eqref{eq:autonomous_left_error} yields the linearized error
dynamics; the detailed derivation is given in the Appendix~\ref{sec:appendix}.
Note that $\bA_{t}$ is time-varying but state-invariant since it depends only on the IMU input.
The LI-EKF prediction step is realized as
\begin{align}
  \frac{d}{dt}\hat{\mathcal{X}}_{t}
   & =
  f_{\bm{u}_{t}}(\hat{\mathcal{X}}_{t})
  \label{eq:nominal_group_dynamics}
  \\
  \frac{d}{dt}\bP_{t}
   & =
  \bA_{t}\,\bP_{t}
  +
  \bP_{t}\,\bA_{t}^{\top}
  +
  \bQ_c
  \\
  \bQ_c
   & =
  \operatorname{diag}
  \bigl(
  \bQ_{\omega},\,
  \bQ_{a},\,
  \bm{0}_{3}
  \bigr)
  \label{eq:cov_prop_continuous}
\end{align}
The covariance $\bP$ represents the uncertainty in the local left-invariant
error coordinates of Eq.~\eqref{eq:left_error_expanded}.

\subsection{Prediction}
\label{ssec:prediction}
\subsubsection{Mean Propagation}
%
By discretizing Eq.~\eqref{eq:continuous_R}--\eqref{eq:continuous_p}, the estimated mean state is propagated in the world frame as:

\begin{align}
  \hat{\mathbf{R}}_{k}^{-}
   & =
  \hat{\mathbf{R}}_{k-1}^{+}
  \operatorname{Exp}
  \left(
  \boldsymbol{\omega}^{\mathrm{imu}}_{k}\Delta t
  \right),
  \label{eq:mean_R}
  \\
  \hat{\mathbf{v}}_{k}^{-}
   & =
  \hat{\mathbf{v}}_{k-1}^{+}
  +
  \left(
  \hat{\mathbf{R}}_{k-1}^{+}\mathbf{a}^{\mathrm{imu}}_{k}
  +
  \mathbf{g}
  \right)\Delta t,
  \label{eq:mean_v}
  \\
  \hat{\mathbf{p}}_{k}^{-}
   & =
  \hat{\mathbf{p}}_{k-1}^{+}
  +
  \hat{\mathbf{v}}_{k-1}^{+}\Delta t
  +
  \frac{1}{2}
  \left(
  \hat{\mathbf{R}}_{k-1}^{+}\mathbf{a}^{\mathrm{imu}}_{k}
  +
  \mathbf{g}
  \right)\Delta t^2,
  \label{eq:mean_p}
\end{align}
with $\Exp$ denoting the $SO(3)$ exponential map and $k \in \mathbb{N}$ is the discrete time index.
\subsubsection{Covariance Propagation}
%
The discrete transition matrix is approximated using a second-order
truncation of the matrix exponential:
\begin{equation}
  \bA_d
  =
  \bI_{9}
  +
  \bA_t \Delta t
  +
  \frac{1}{2}
  \left(
  \bA_t
  \right)^2
  \Delta t^2
  \label{eq:Ad_left}
\end{equation}
The quadratic term retains the second-order coupling among local
orientation, velocity, and position errors. In particular,
$-\skewmat{\bm{a}^{\mathrm{imu}}}$ transfers orientation uncertainty into velocity,
whereas $\bI_3$ transfers velocity uncertainty into position. The covariance is propagated as:
\begin{equation}
  \bP^{-}_{k}
  =
  \bA_d
  \bP^{+}_{k-1}
  \bA_d^{\top}
  +
  \bQ_d
  \label{eq:cov_prop_left}
\end{equation}
with $\bQ_d =\bQ_c\Delta t$,  the first-order discretization of the continuous noise density $\bQ_c$.

\subsection{Velocity Update}
\label{ssec:kin_update}

Following~\cite{invariantekf2017}, a measurement model is \emph{left-invariant} if and only if:
\begin{equation}
  \bm{Y}_{t}
  =
  \mathcal{X}_{t}\,\bm{b}
  +
  \bm{V}_{t}
  \label{eq:left_invariant_measurement}
\end{equation}
Measurements expressed in the world frame of reference admit the left-invariant
form~\eqref{eq:left_invariant_measurement}.  To this end, the kinematic base-velocity measurement of~\eqref{eq:kin_vel_measurement} can be written as:
\begin{equation}
  \underbrace{
    \begin{pmatrix}
      \bv^{\rm{kin}}_t \\
      1                \\
      0
    \end{pmatrix}
  }_{\bm{Y}_{t}}
  =
  \underbrace{
    \begin{pmatrix}
      \bR_{t}           & \bv_{t} & \bp_{t} \\
      \bm{0}_{1\times3} & 1       & 0       \\
      \bm{0}_{1\times3} & 0       & 1
    \end{pmatrix}
  }_{\mathcal{X}_{t}}
  \underbrace{
    \begin{pmatrix}
      \bm{0}_{3\times 1} \\
      1                  \\
      0
    \end{pmatrix}
  }_{\bm{b}}
  +
  \underbrace{
    \begin{pmatrix}
      \bm{n}_{v} \\
      0          \\
      0
    \end{pmatrix}
  }_{\bm{V}_{t}}
  \label{eq:world_velocity_measurement}
\end{equation}

\paragraph{Realizing the left-invariant update}
For a left-invariant observation, the corrected estimate is obtained by
right multiplication~\cite{hartley2018contact}:
\begin{equation}
  \hat{\mathcal{X}}_{t}^{+}
  =
  \hat{\mathcal{X}}_{t}\,
  \Exp\!
  \Bigl(
  \bL_{t}
  \bigl(
    \hat{\mathcal{X}}_{t}^{-1}\bm{Y}_{t}
    -
    \bm{b}
    \bigr)
  \Bigr)
  \label{eq:left_invariant_update}
\end{equation}
Substituting~\eqref{eq:left_invariant_measurement} into~\eqref{eq:left_invariant_update}
gives
\begin{equation}
  \hat{\mathcal{X}}_{t}^{+}
  =
  \hat{\mathcal{X}}_{t}\,
  \Exp\!
  \Bigl(
  \bL_{t}
  \bigl(
    \hat{\mathcal{X}}_{t}^{-1}\mathcal{X}_{t}\,\bm{b}
    +
    \hat{\mathcal{X}}_{t}^{-1}\bm{V}_{t}
    -
    \bm{b}
    \bigr)
  \Bigr)
  \label{eq:left_update_expanded}
\end{equation}
where $\bL_{t} \in \mathbb{R}^{9\times5}$ is a linear map that takes the innovation into the Lie algebra of $\SEtwo$.
Since the left-invariant error satisfies
$\bm{\eta}_{t}^{L}=\mathcal{X}_{t}^{-1}\hat{\mathcal{X}}_{t}$
and therefore
$\bigl(\bm{\eta}_{t}^{L}\bigr)^{-1}
  =
  \hat{\mathcal{X}}_{t}^{-1}\mathcal{X}_{t}
  =
  \Exp(-\bm{\xi}_{t})$,
the updated error evolves as:
\begin{equation}
  \bm{\eta}_{t}^{L+}
  =
  \mathcal{X}_{t}^{-1}\hat{\mathcal{X}}_{t}^{+}
  =
  \bm{\eta}_{t}^{L}\,
  \Exp\!
  \Bigl(
  \bL_{t}
  \bigl(
    \bigl(\bm{\eta}_{t}^{L}\bigr)^{-1}\bm{b}
    -
    \bm{b}
    +
    \hat{\mathcal{X}}_{t}^{-1}\bm{V}_{t}
    \bigr)
  \Bigr)
  \label{eq:left_error_update}
\end{equation}
As in~\cite{hartley2018contact}, the homogeneous innovation is reduced by the selector:
\begin{equation}
  \bm{\Pi}
  =
  \begin{pmatrix}
    \bI_{3} & \bm{0}_{3\times2}
  \end{pmatrix}
  \label{eq:pi_selector}
\end{equation}
so that,
\begin{equation}
  \bL_{t}
  \bigl(
  \hat{\mathcal{X}}_{t}^{-1}\bm{Y}_{t}
  -
  \bm{b}
  \bigr)
  =
  \bK_{t}\,
  \bm{\Pi}\,
  \hat{\mathcal{X}}_{t}^{-1}\bm{Y}_{t}
  \label{eq:gain_reduction}
\end{equation}
Using the first-order approximation
$\bigl(\bm{\eta}_{t}^{L}\bigr)^{-1}
  =
  \Exp(-\bm{\xi}_{t}^{\wedge})
  \simeq
  \bm{I}-\bm{\xi}_{t}^{\wedge}$
and the small-angle expansion of the exponential map:
\begin{align}
  \bm{I}+\bm{\xi}_{t}^{+\wedge}
   & \simeq
  \bigl(\bm{I}+\bm{\xi}_{t}^{\wedge}\bigr)
  \Exp\!
  \Bigl(
  \bL_{t}
  \bigl(
    -\bm{\xi}_{t}^{\wedge}\bm{b}
    +
    \hat{\mathcal{X}}_{t}^{-1}\bm{V}_{t}
    \bigr)
  \Bigr)
  \nonumber \\
   & \simeq
  \bigl(\bm{I}+\bm{\xi}_{t}^{\wedge}\bigr)
  \Bigl(
  \bm{I}
  +
  \bL_{t}
  \bigl(
    -\bm{\xi}_{t}^{\wedge}\bm{b}
    +
    \hat{\mathcal{X}}_{t}^{-1}\bm{V}_{t}
    \bigr)^{\wedge}
  \Bigr)
  \label{eq:error_linearization}
\end{align}
Neglecting second-order terms yields:
\begin{equation}
  \bm{\xi}_{t}^{+\wedge}
  \simeq
  \bm{\xi}_{t}^{\wedge}
  +
  \Bigl(\bK_{t}\,
  \bm{\Pi}
  \bigl(
  -\bm{\xi}_{t}^{\wedge}\bm{b}
  +
  \hat{\mathcal{X}}_{t}^{-1}\bm{V}_{t}
  \bigr)\Bigr)^{\wedge}
  \label{eq:xi_wedge_update}
\end{equation}
\paragraph{Velocity measurement linearization}
For the world-velocity observation~\eqref{eq:world_velocity_measurement},
$\bm{b}=( \bm{0}_{3\times1},\,1,\,0)^{\top}$ extracts the
velocity error component,
\begin{equation}
  -\bm{\xi}_{t}^{\wedge}\bm{b}_{v}
  =
  \begin{pmatrix}
    -\bm{\xi}_{v} \\
    0             \\
    0
  \end{pmatrix}
  \label{eq:xi_b_velocity}
\end{equation}
while the noise term transforms as:
\begin{equation*}
  \hat{\mathcal{X}}_{t}^{-1}\bm{V}_{t}
  =
  \begin{pmatrix}
    \hat{\bR}_{t}^{\top}
                      &
    -\hat{\bR}_{t}^{\top}\hat{\bv}_{t}
                      &
    -\hat{\bR}_{t}^{\top}\hat{\bp}_{t}
    \\
    \bm{0}_{1\times3} & 1 & 0 \\
    \bm{0}_{1\times3} & 0 & 1
  \end{pmatrix}
  \begin{pmatrix}
    \bm{n}_{v} \\
    0          \\
    0
  \end{pmatrix}
  =
  \begin{pmatrix}
    \hat{\bR}_{t}^{\top}\bm{n}_{v} \\
    0                              \\
    0
  \end{pmatrix}
  \label{eq:noise_transform_velocity}
\end{equation*}
Collecting terms in vector form with the vee operation gives the linearized update:
\begin{equation}
  \bm{\xi}_{t}^{+}
  =
  \bm{\xi}_{t}
  -
  \bK_{t}
  \Bigl(
  \bH_{v}\,\bm{\xi}_{t}
  -
  \hat{\bR}_{t}^{\top}\bm{n}_{v}
  \Bigr)
  \label{eq:velocity_linear_update}
\end{equation}
with observation matrix and base-frame measurement covariance:
\begin{equation}
  \bH_{v}
  =
  \begin{pmatrix}
    \bm{0}_{3} & \bI_{3} & \bm{0}_{3}
  \end{pmatrix},
  \qquad
  \bN_{v}
  =
  \hat{\bR}_{t}^{\top}\,
  \tfrac{\bm{\Sigma}^{\rm{kin}}_t}{\Delta t}\,
  \hat{\bR}_{t}
  \label{eq:H_N_velocity}
\end{equation}

The innovation covariance and Kalman gain can be computed as:
\begin{align}
  \bS_v
   & =
  \bN_v
  +
  \bH_v\bP^{-}\bH_v^{\top}
  \label{eq:vel_innov_cov}
  \\
  \bK_v
   & =
  \bP^{-}\bH_v^{\top}\bS_v^{-1} \\
  \delta\bm{\xi}^v
   & =
  \bK_{v}\bm \hat{\bm{R}}_t^\top \left(\bm{v}_t^{\rm{kin}} - \bm{\hat{v}}_t\right)
  \label{eq:vel_gain}
\end{align}
here, $\delta\bm{\xi}^{v} \in \mathbb{R}^{9}$ denotes the correction generated by the velocity measurement.
The Joseph-form covariance update is:
\begin{align}
  \bP^{+}
   & =
  (\bI-\bK_v\bH_v)
  \bP^{-}
  (\bI-\bK_v\bH_v)^{\top}
  \nonumber \\
   & \quad
  +
  \bK_v\bN_v\bK_v^{\top}
  \label{eq:vel_cov}
\end{align}
The velocity measurement constrains the instantaneous base velocity and
thereby prevents base velocity errors from continuously integrating into
base position. However, it does not directly determine an accumulated global
base position offset.

\subsection{Incorporating the IMU Biases}
\label{ssec:imubiases}
Augmenting the filter state as $(\mathcal{X}_t, \bm{b}_t) \in \SEtwo \times \R^{6}$, where
$\bm{b}_t = (\bm{b}^{g}_t, \bm{b}^{a}_t)$ denotes the gyroscope and accelerometer biases, breaks the group-affine
property, as no group operation on the augmented space satisfies it.
Similarly to~\cite{invariantekf2017} where the ``imperfect'' right-invariant EKF is demonstrated to outperform standard EKFs although
the theoretical properties of invariant EKF cannot be guaranteed. Experimentally, we've observed the same benefits for our left-invariant formulation.

Modeling the bias state as zero-mean Gaussian processes:
\begin{align}
  \frac{d}{dt} \bm{b}^g_t & = \bm{w}_t^{b^g} \qquad
  \bm{w}_t^{b^g}\sim\cN\!\bigl(\bm{0},\,\bQ_{b^g}\bigr) \\
  \frac{d}{dt} \bm{b}^a_t & = \bm{w}_t^{b^a} \qquad
  \bm{w}_t^{b^a}\sim\cN\!\bigl(\bm{0},\,\bQ_{b^a}\bigr)
\end{align}
with the input now being:

\begin{align}
  \bm{\omega}_{t}^{\mathrm{imu}}
   & =
  \tilde{\bm{\omega}}_{t}
  +
  \bm{b}^g_t
  +
  \bm{w}_t^{g}
  \\
  \bm{a}_{t}^{\mathrm{imu}}
   & =
  \tilde{\bm{a}}_{t}
  +
  \bm{b}^a_t
  +
  \bm{w}_t^{a}
\end{align}
The augmented left-invariant error is given by:
\begin{align*}
  \bm{e}_t^L=\left(\mathcal{X}_t^{-1}\hat{\mathcal{X}_t}, \bm{b}_t - \hat{\bm{b}}_t\right) = \left(\bm{\eta}_{t}^{L}, \bm{\zeta}_t\right) =\left(\bm{\eta}_{t}^{L}, \begin{bmatrix} \bm{\zeta}_t^{b^g} \\ \bm{\zeta}_t^{b^a}\end{bmatrix} \right)
\end{align*}
and the continuous-time dynamics now become:
\begin{align}
  \frac{d}{dt}\bR_{t}
   & =
  \bR_{t}\,
  \skewmat{
  \bm{\omega}_{t}^{\mathrm{imu}}
  -
  \bm{b}_t^{g}
  -
  \bm{w}_t^{g}
  }
  \\
  \frac{d}{dt}\bv_{t}
   & =
  \bR_{t}
  \bigl(
  \bm{a}_{t}^{\mathrm{imu}}
  -
  \bm{b}_t^{a}
  -
  \bm{w}_t^{a}
  \bigr)
  +
  \bg
  \\
  \frac{d}{dt}\bp_{t}
   & =
  \bv_{t}
\end{align}
To this end, the augmented left-invariant error dynamics are:
\begin{align}
  \label{eq:augmented_left_error_dynamics}
  \frac{d}{dt}\bm{e}_t^L = \left(\frac{d}{dt} \bm{\eta}_t^{L}, \frac{d}{dt} \bm{\zeta}_t \right)
\end{align}
Linearizing~\eqref{eq:augmented_left_error_dynamics} similarly to Appendix~\ref{sec:appendix}, we get:
\begin{align*}
  \frac{d}{dt} \left(\bR_t^\top \hat{\bR}_t\right)                         & = -\left(\bm{\omega}^{\rm{imu}} \right)_{\times} \bm{\xi}^{\bm{R}}_t - \bm{\zeta}^{b^g}_t  + \bm{w}_t^{g} \\
  \frac{d}{dt} \left(\bR_t^\top \left(\hat{\bm{v}} - \bm{v}\right)\right) & = -\left(\bm{a}^{\rm{imu}} \right)_{\times} \bm{\xi}^{\bm{R}}_t -
  \left(\bm{\omega}^{\rm{imu}} \right)_{\times} \bm{\xi}^{\bm{v}}_t - \bm{\zeta}^{b^a}_t + \bm{w}_t^{a}                                                                                \\
  \frac{d}{dt} \left(\bR_t^\top \left(\hat{\bm{p}} - \bm{p}\right)\right) & = \bm{\xi}^{\bm{v}}_t - \left(\bm{a}^{\rm{imu}} \right)_{\times} \bm{\xi}^{\bm{p}}_t                      \\
\end{align*}
This can be written as:
\begin{equation}
  \frac{d}{dt}\bm{e}_t = \bm{A}_t \bm{e}_t + \bm{w}_t
\end{equation}
with:
\begin{align}
  \bm{A}_t & =
  \begin{pmatrix}
    -\skewmat{\bm{\omega}_{t}^{\mathrm{imu}}}
     & \bm{0}_{3}
     & \bm{0}_{3}
     & -\bm{I}_{3}
     & \bm{0}_{3}
    \\
    -\skewmat{\bm{a}_{t}^{\mathrm{imu}}}
     & -\skewmat{\bm{\omega}_{t}^{\mathrm{imu}}}
     & \bm{0}_{3}
     & \bm{0}_{3}
     & -\bm{I}_{3}
    \\
    \bm{0}_{3}
     & \bI_{3}
     & -\skewmat{\bm{\omega}_{t}^{\mathrm{imu}}}
     & \bm{0}_{3}
     & \bm{0}_{3}
    \\
    \bm{0}_{3}
     & \bm{0}_{3}
     & \bm{0}_{3}
     & \bm{0}_{3}
     & \bm{0}_{3}
    \\
    \bm{0}_{3}
     & \bm{0}_{3}
     & \bm{0}_{3}
     & \bm{0}_{3}
     & \bm{0}_{3}
  \end{pmatrix}
\end{align}
and since:
\begin{align}
  \bm{w}_t & = \left(\bm{w}_t^g, \bm{w}^a_t, \bm{0}_{3\times1}, \bm{w}_t^{b^g}, \bm{w}_t^{b^a} \right)
\end{align}
the noise matrix becomes:
\begin{align}
  \bm{Q}_c = diag\left(\bm{Q}_\omega, \bm{Q}_a, \bm{0}_{3}, \bm{Q}_{b^g}, \bm{Q}_{b^a}\right)
\end{align}
which is the model used in the prediction step in Section~\ref{ssec:prediction}. For velocity update (Section~\ref{ssec:kin_update}) the observation matrix $\bm{H}_v$ becomes:
\begin{equation}
  \bm{H}_v = \begin{pmatrix} \bm{0}_3 & \bm{I}_3 & \bm{0}_3 & \bm{0}_3 & \bm{0}_3 \end{pmatrix}
\end{equation}
The key difference here is that each update step instead of computing a $\delta\bm{\xi}$ to correct the base state estimate with, we compute a $\delta{\bm{e}} = \left(\delta\bm{\xi}, \delta \bm{\zeta}\right)$ that corrects the base state and the biases state respectively.
\subsection{State Retraction}
\label{ssec:retraction}

After each measurement update, the estimated correction
\begin{align}
  \delta\bm{\xi}
                    & =
  \begin{pmatrix}
    \delta\bm{\xi}_R^{\top}
     &
    \delta\bm{\xi}_v^{\top}
     &
    \delta\bm{\xi}_p^{\top}
  \end{pmatrix}^{\top} \\
  \delta \bm{\zeta} & =
  \begin{pmatrix}
    \delta\bm b_g^{\top}
     &
    \delta\bm b_a^{\top}
  \end{pmatrix}^{\top}
\end{align}
is retracted onto the state manifold. Because the state estimation error is
left-invariant, the correction is applied to the estimated group state from the right, this translates to:
\begin{align}
  \hat{\bR}^{+}
   & =
  \hat{\bR}
  \Exp\left(
  \delta\bm{\xi}_R
  \right)
  \label{eq:retract_R}
  \\
  \hat{\bv}^{+}
   & =
  \hat{\bv}
  +
  \hat{\bR}
  \delta\bm{\xi}_v
  \label{eq:retract_v}
  \\
  \hat{\bp}^{+}
   & =
  \hat{\bp}
  +
  \hat{\bR}
  \delta\bm{\xi}_p
  \label{eq:retract_p}
\end{align}
The linear updates in \eqref{eq:retract_v}--\eqref{eq:retract_p} correspond to the first-order
approximation of the full $\SEtwo$ exponential.
While the IMU biases are linearly updated as:
\begin{align}
  \hat{\bm b}_{g}^{+}
   & =
  \hat{\bm b}_{g}
  +
  \delta\bm b_g
  \label{eq:retract_bg}
  \\
  \hat{\bm b}_{a}^{+}
   & =
  \hat{\bm b}_{a}
  +
  \delta\bm b_a
\end{align}

\section{Simultaneous Base and Elevation Estimation}
\label{sec:coupling}

\begin{algorithm}[t]
  \scriptsize
  \caption{Simultaneous Base and Elevation Estimation (SBEE)}
  \label{alg:update}
  \begin{algorithmic}[1]

    \REQUIRE Prior state
    $(\hat{\bR},\hat{\bv},\hat{\bp},
      \hat{\bm b}_g,\hat{\bm b}_a)$ and covariance $\bP$
    \REQUIRE IMU measurements
    $(\bm{\omega}^{\mathrm{imu}},
      \bm a^{\mathrm{imu}})$
    \REQUIRE Foot kinematics and contact confidence
    $\{{}^{b}\bp_f,{}^{b}\dot{\bp}_f,CC_f\}$
    \REQUIRE Terrain map $\mathcal{M}$

    \STATE \textbf{Phase 1: LI-EKF inertial prediction}
    \STATE Bias-correct the IMU input:
    $\bm{\omega}^{\mathrm{imu}}\!-\!\hat{\bm b}_g$,
    $\bm a^{\mathrm{imu}}\!-\!\hat{\bm b}_a$
    \STATE Propagate $(\hat{\bR},\hat{\bv},\hat{\bp})$
    using \eqref{eq:mean_R}--\eqref{eq:mean_p};
    hold $(\hat{\bm b}_g,\hat{\bm b}_a)$ constant
    \STATE Propagate $\bP$ using
    \eqref{eq:Ad_left}--\eqref{eq:cov_prop_left}

    \STATE \textbf{Phase 2: Proprioceptive velocity update}
    \STATE Compute ${}^{w}\bv_b^{\mathrm{kin}}$ and
    ${}^{w}\bm{\Sigma}_v^{\mathrm{kin}}$ using
    \eqref{eq:kin_vel_measurement}--\eqref{eq:kin_vel_covariance}

    \IF{$\sum_f CC_f> 0$}
    \STATE Form $\bH_v$ and $\bN_v$ using
    \eqref{eq:H_N_velocity}
    \STATE Compute $\delta\bm{\xi}^v$ and $\bP^{+}$ using
    \eqref{eq:vel_innov_cov}--\eqref{eq:vel_cov}
    \STATE Retract using
    \eqref{eq:retract_R}--\eqref{eq:retract_p}
    and \eqref{eq:retract_bg}
    \ENDIF

    \STATE \textbf{Phase 3: Terrain-map update
      ($\mathcal{M}\leftarrow$ LI-EKF)}
    \STATE Filter the contact confidences

    \FORALL{$f$ such that $CC_f>CC_{\min}$}
    \STATE ${}^{w}\hat{\bp}_f
      \leftarrow {}^{w}\hat{\bR}_{b} {}^{b}\bp_f + {}^{w}\hat{\bp}_{b}$
    {\normalfont\texttt{//}~\eqref{eq:foot_world}}
    \STATE Compute $\sigma_{z,f}^{2}$ using
    \eqref{eq:meas_var}

    \IF{robot has flat feet \AND contact is stable}
    \STATE Extract the surface normal $\bn_f$ from the foot orientation
    \ELSIF{robot has point feet}
    \STATE $\bn_f \leftarrow {}^{w}\bR_{b}\,\bm{e_z}$
    \ELSE
    \STATE $\bn_f \leftarrow \emptyset$ \COMMENT{no terrain update}
    \ENDIF
    \STATE $\bm{\ell}_f\leftarrow
      ({}^{w}\hat p_f^x,{}^{w}\hat p_f^y)^\top$,
    $z_f\leftarrow{}^{w}\hat p_f^z$
    \STATE $\mathcal{M}.\textsc{update}
      (\bm{\ell}_f,z_f,\sigma_{z,f}^{2},\bn_f)$
    {\normalfont\texttt{//}~\eqref{eq:kalman_gain}--\eqref{eq:var_update}}
    \STATE $\mathcal{M}.\textsc{propagateNeighborhood}
      (\bm{\ell}_f,\bn_f)$
    {\normalfont\texttt{//}~\eqref{eq:plane_proj}--\eqref{eq:dist_var}}
    \ENDFOR

    \STATE \textbf{Phase 4: Map maintenance}

    \IF{$\lVert\hat{\bp}^{xy}-\bm o\rVert_2>d_{\max}$}
    \STATE $\mathcal{M}.\textsc{recenter}
      (\hat{\bp}^{xy})$
    \ENDIF

    \STATE $\mathcal{M}.
      \textsc{interpolateContactPoints}()$
    {\normalfont\texttt{//}~\eqref{eq:idw}}

    \STATE \textbf{Phase 5: Terrain correction
      (LI-EKF $\leftarrow\mathcal{M}$)}

    \FORALL{stable contacts $f$ such that $CC_f\geq CC_{stable}$}
    \STATE $\bm{\ell}_f\leftarrow
      ({}^{w}\hat p_f^x,{}^{w}\hat p_f^y)^\top$
    {\normalfont\texttt{//}~\eqref{eq:terrain_query_location}}
    \STATE $(h_f,\sigma_f^2)\leftarrow
      \mathcal{M}.\textsc{getElevation}(\bm{\ell}_f)$

    \IF{a valid elevation is available}
    \STATE Compute the scalar innovation $z_{h,f}$ using
    \eqref{eq:terrain_height_discrepancy}
    \STATE Form $\mathbf{H}_h$ using \eqref{eq:H_terrain}

    \STATE Compute $\delta\mathbf{e}^{h}_{f}$ and $\mathbf{P}^{+}$ using \eqref{eq:K_terrain}--\eqref{eq:P_terrain}
    \STATE Retract using
    \eqref{eq:retract_R}--\eqref{eq:retract_p}
    and \eqref{eq:retract_bg}
    \ENDIF
    \ENDFOR

    \RETURN Updated state
    $(\hat{\bR}^{+},\hat{\bv}^{+},\hat{\bp}^{+},
      \hat{\bm b}_{g}^{+},\hat{\bm b}_{a}^{+})$,
    covariance $\bP^{+}$ and terrain map $\mathcal{M}$
  \end{algorithmic}
\end{algorithm}
The terrain elevation mapper of Section~\ref{sec:terrain_map} and the LI-EKF floating-base estimation of Section~\ref{sec:liekf} operate in a bidirectional estimation
loop that constitute the proposed Simultaneous Base and Elevation Estimation (SBEE) framework illustrated in Fig.~\ref{fig:flowchart}.

The two components constantly exchange information with one another. In the forward direction, the current base state estimate
registers each contact in the world frame and inserts it to the local elevation map as a height measurement weighted by contact uncertainty.
In the backward direction, whenever a stable contact lands on a cell that already carries a converged elevation, the stored height
is converted into an uncertainty-weighted vertical world pseudo measurement
that anchors the current base state.

Note that the map is deliberately not part of the filter state: it acts as an external, uncertainty-aware
memory of the terrain the robot has already stepped on, and its influence on the filter is confined to the
vertical base position. The strength of that influence is governed by the variance of the queried cell, so
freshly observed cells correct the base state weakly, whereas repeatedly contacted areas act as a strong
anchor, as detailed in Section~\ref{ssec:ekf_terrain} below.

\subsection{Terrain-Elevation Update}
\label{ssec:ekf_terrain}

The terrain map provides a persistent world frame height reference. It does not create an external
absolute-altitude measurement; rather, it constrains vertical drift relative to the initialized world frame.
Since this is a scalar measurement update, and since the contact location is not explicitly part of the state,
the derivation of Section~\ref{ssec:kin_update} cannot be followed exactly. While one could reconstruct the
three-dimensional base position by assuming a terrain-anchored leg contact and projecting the resulting update
onto the vertical axis, this approach requires transforming world frame spatial uncertainty into the local base
frame. Artificially inflating the world $x$- and $y$-uncertainties causes non-physical uncertainty leakage into
the local frame upon rotation, whereas constraining them to the current filter covariance causes filter
overconfidence. We overcome these issues by formulating a scalar update directly from the terrain measurement.
As the correction is applied to the bias-augmented state of Section~\ref{ssec:imubiases}, it is stated here
directly in the augmented error coordinates
$\delta\bm{e} = (\delta\bm{\xi}, \delta\bm{\zeta}) \in \mathbb{R}^{15}$.

For a contact $f$, the predicted world frame contact position
${}^{w}\hat{\bp}_{f}$ is obtained by evaluating Eq.~\eqref{eq:foot_world}
at the current estimate $(\hat{\bR},\hat{\bp})$.
Since the map is indexed by horizontal position, the vertical component of ${}^{w}\hat{\bp}_{f}$ is discarded and
the remaining components address the cell
\begin{equation}
  \bm{\ell}_f
  =
  \begin{pmatrix}
    {}^{w}\hat p_f^x \\
    {}^{w}\hat p_f^y
  \end{pmatrix}
  \label{eq:terrain_query_location}
\end{equation}
whose stored elevation state $(h_f,\sigma_f^2)$ is retrieved through the indexing of
Section~\ref{ssec:map_repr}. The scalar base-height innovation is
\begin{equation}
  z_{h,f}
  =
  h_f
  -
  \bm{e}_3^{\top}\,{}^{w}\hat{\bp}_f
  \approx
  \bm{e}_3^{\top} \hat{\bR}\, \bm{\xi}_p
  \label{eq:terrain_height_discrepancy}
\end{equation}
where $\bm{\xi}_p$ is the position block of the estimation error. Consequently, the terrain observation matrix is
\begin{equation}
  \bH_h
  =
  \begin{pmatrix}
    \bm{0}_{1\times3} & \bm{0}_{1\times3} & \bm{e}_3^{\top}\hat{\bR} &
    \bm{0}_{1\times3} & \bm{0}_{1\times3}
  \end{pmatrix}
  \label{eq:H_terrain}
\end{equation}
so that $z_{h,f} = \bH_h\,\delta\bm{e} + n_{h,f}$, where
$n_{h,f}\sim\mathcal{N}\!\left(0,\sigma_f^{2}\right)$ is the noise of the
terrain pseudo-measurement, whose variance is the stored variance of the
queried cell retrieved alongside $h_f$. A contact is considered stable, and
therefore eligible to correct the base state, when its confidence exceeds a
threshold $CC_{\mathrm{stable}}\in(0,1]$ whichis deliberately
stricter than the one that admits a contact into the map, $CC_{\min}$, so
that the filter is only anchored by footholds that are confidently loaded.

For every stable contact satisfying $CC_f \geq CC_{\mathrm{stable}}$, a scalar update is applied sequentially.
Because the map is not part of the filter state and its cells may carry errors not captured by $\sigma_f^2$,
the terrain innovation is not allowed to act on every state. Following the Schmidt--Kalman partial
update~\cite{brink2017partial}, the gain is restricted by the selector
\begin{equation}
  \mathbf{\Lambda}_{h} = \operatorname{blkdiag}\!\left(\mathbf{0}_{3},\, \mathbf{0}_{3},\,
  \hat{\mathbf{R}}^{\top}\mathbf{e}_{3}\mathbf{e}_{3}^{\top}\hat{\mathbf{R}},\,
  \mathbf{0}_{3},\, \mathbf{0}_{3}\right),
  \label{eq:L_terrain}
\end{equation}
which keeps only the world-vertical component of the position error, so that
\begin{align}
  \mathbf{K}_{h,f} &= \frac{\mathbf{\Lambda}_{h}\mathbf{P}^{-}\mathbf{H}_{h}^{\top}}
                           {\sigma_{f}^{2} + \mathbf{H}_{h}\mathbf{P}^{-}\mathbf{H}_{h}^{\top}}
  \label{eq:K_terrain} \\
  \delta\mathbf{e}^{h}_{f} &= \mathbf{K}_{h,f}\, z_{h,f}
  \label{eq:dx_terrain} \\
  \mathbf{P}^{+} &= (\mathbf{I}-\mathbf{K}_{h,f}\mathbf{H}_{h})\,\mathbf{P}^{-}\,
                   (\mathbf{I}-\mathbf{K}_{h,f}\mathbf{H}_{h})^{\top}
                   + \mathbf{K}_{h,f}\,\sigma_{f}^{2}\,\mathbf{K}_{h,f}^{\top}
  \label{eq:P_terrain}
\end{align}
each correction being retracted onto the state manifold before the next contact is processed. The correction
is thus a pure vertical translation of the base: velocity, orientation and IMU biases, and every block of
$\mathbf{P}$ not involving the position error, are left unchanged (Table~\ref{tab:terrain_ablation}). Since
$\mathbf{K}_{h,f}$ is no longer the minimum-variance gain, the Joseph form~\eqref{eq:P_terrain} is required to
keep the covariance consistent; the scalar innovation covariance still requires no matrix inversion.

The magnitude of the correction is governed entirely by the availability and the uncertainty of the queried cell.
If no valid elevation exists at $\bm{\ell}_f$, the update is omitted and the estimator continues to rely on IMU
propagation and kinematic velocity corrections alone. A cell that has received only a few observations retains a
comparatively large $\sigma_f^2$ and therefore yields a weak correction, which prevents the base state from
following an uncertain map estimate. Conversely, repeated observations drive $\sigma_f^2$ toward the prescribed
minimum variance, so that contacting a well-mapped location produces a large Kalman gain and strongly reduces the
height inconsistency accumulated relative to the persistent map. This last regime constitutes the proposed local,
touch-based loop-closure-like correction.

\subsection{Per-Cycle Execution}
At each cycle, the IMU first propagates the LI-EKF. The proprioceptive world frame velocity correction is then applied, using the weighted contact confidence
kinematic measurement of Eqs.~\eqref{eq:kin_vel_measurement}--\eqref{eq:kin_vel_covariance}. The corrected base state is then used to insert stable contact observations into the terrain map.
After the interpolation and maintenance of the map, a vertical pseudo-position measurement is provided. The map is always updated with the best available estimate
and the terrain correction is always computed against a map that reflects the current footholds. A detailed description of this process is outlined in Algorithm~\ref{alg:update}.

\subsection{Terrain Map as External Contact Memory}

A defining property of the proposed formulation is that the dimension of the filter state does not depend on the morphology of the robot or on the number
of active contacts. Instead of augmenting the state with a contact model per leg, the LI-EKF consumes the leg kinematics as a single 3-D world frame velocity
measurement, while the terrain map is utilized as contact memory. The state therefore remains $\mathrm{SE}_2(3)\times\mathbb{R}^{6}$ throughout, which
means that the proposed framework can be utilized for a robot with an arbitrary number of legs without changing the formulation. Since foot geometry
is incorporated only through the measurement, point feet contribute a single height sample per contact whereas flat feet additionally provide a
surface normal that propagates over the local tangent plane via Eqs.~\eqref{eq:plane_proj}--\eqref{eq:dist_var}.

\section{Results}
\label{sec:results}

This section presents the experimental evaluation of the proposed SBEE framework.
The evaluation is conducted in two stages.
First, simulated experiments in MuJoCo~\cite{mujoco}, on a bipedal and a quadrupedal
platform and challenging terrains are used to assess the contribution of the terrain-elevation update
to estimator robustness. Second, SBEE is compared with three state-of-the-art proprioceptive
state estimators on the ANYmal-D GrandTour Dataset~\cite{grandtour2026}. For each experiment,
we first showcase qualitative analysis followed by quantitative evaluation of the estimator performance.
For consistency, we reproduce the evaluation protocol and the reference results reported
in the recent benchmarking study of~\cite{nistico2026benchmark}.

\subsection{Evaluation Metrics}
\label{ssec:metrics}

All configurations were evaluated over the time interval common to the ground truth with
the ground truth interpolated at the estimator timestamps. The simulator ground truth was expressed in the
estimator's initial reference frame by removing the initial position and the initial yaw angle only, so that the
inclination of the terrain remains physical along the vertical axis. No subsequent trajectory alignment was
applied, since aligning each estimate independently would absorb part of the very drift that the ablation is
intended to expose.

For the position and linear velocity, the vector root-mean-square error was calculated as
\begin{equation}
  \operatorname{RMSE}_{\mathbf{x}}
  =
  \sqrt{
  \frac{1}{N}
  \sum_{i=1}^{N}
  \left\|
  \hat{\mathbf{x}}_{i}-\mathbf{x}_{i}
  \right\|_{2}^{2}
  },
  \qquad
  \mathbf{x}
  \in
  \left\{
  {}^{w}\mathbf{p}_{b},
  {}^{w}\mathbf{v}_{b}
  \right\},
  \label{eq:vector_rmse}
\end{equation}
where $\mathbf{x}_{i}$ and $\hat{\mathbf{x}}_{i}$ denote the ground truth and estimated vectors at the $i$-th
timestamp, respectively, and $N$ is the number of evaluated samples. Since the terrain-elevation update acts on
the vertical axis only, the horizontal and vertical components of the position error are additionally reported
in isolation. $\operatorname{RMSE}_{xy}$ is computed as in~\eqref{eq:vector_rmse} over the horizontal components
$({}^{w}p_{b}^{x},{}^{w}p_{b}^{y})$ alone. Let $e_{i} = \mathbf{e}_{3}^{\top}(\hat{\mathbf{x}}_{i}-\mathbf{x}_{i})$,
with $\mathbf{x} \in \{{}^{w}\mathbf{p}_{b}, {}^{w}\mathbf{v}_{b}\}$, denote the vertical error at the $i$-th
timestamp. We report its root-mean-square value and its mean,
\begin{equation}
  \operatorname{RMSE}_{z}
  =
  \sqrt{\frac{1}{N}\sum_{i=1}^{N} e_{i}^{2}},
  \qquad
  \bar{e}_{z}
  =
  \frac{1}{N}\sum_{i=1}^{N} e_{i},
  \label{eq:vertical_error}
\end{equation}
where $\bar{e}_{z}$ isolates the vertical offset accumulated by the estimator during the experiment. Relative
improvements reported for $\bar{e}_{z}$ are computed on magnitudes, as the quantity is signed.

The orientation error was calculated using the geodesic distance on $\mathrm{SO}(3)$. The relative rotation error
at the $i$-th timestamp is given by
\begin{equation}
  \Delta\mathbf{R}_{i}
  =
  \left(
  {}^{w}\mathbf{R}_{b,i}
  \right)^{\top}
  {}^{w}\hat{\mathbf{R}}_{b,i},
  \label{eq:relative_rotation_error}
\end{equation}
and its angular magnitude is
\begin{equation}
  \theta_i
  =
  \cos^{-1}
  \left(
  \frac{
    \operatorname{tr}
    \left(
    \Delta\mathbf{R}_{i}
    \right)-1
  }{2}
  \right)
  \label{eq:orientation_geodesic_error}
\end{equation}
The orientation RMSE, reported in degrees, is therefore
\begin{equation}
  \operatorname{RMSE}_{R}
  =
  \frac{180}{\pi}
  \sqrt{
    \frac{1}{N}
    \sum_{i=1}^{N}
    \theta_i^{2}
  }
  \label{eq:orientation_rmse}
\end{equation}

The metrics above are used throughout the simulated experiments. The comparative evaluation of
Section~\ref{ssec:anymal_grandtour} instead follows the protocol of the reference benchmark~\cite{nistico2026benchmark}, reporting
absolute trajectory error and relative pose error computed with the \texttt{evo} package after rigid
$\mathrm{SO}(3)$ alignment of the estimated trajectory. Owing to that alignment, the two sets of figures are not
directly comparable.

\begin{figure}[b]
  \centering
  \subfloat[Rough terrain.\label{fig:rough_experiment}]{%
    \includegraphics[width=0.49\columnwidth]{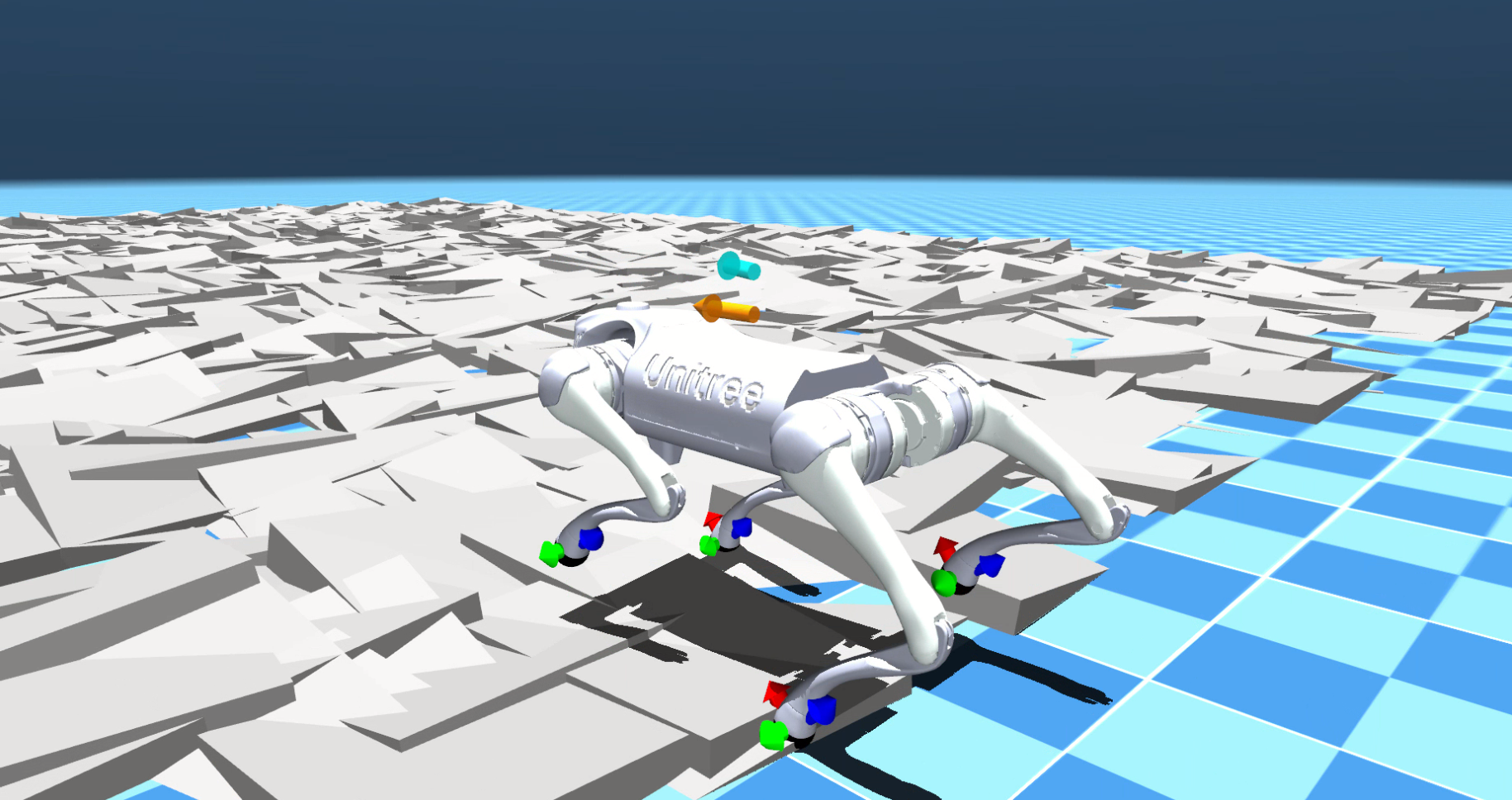}}\hfil
  \subfloat[Sloped terrain, $10^\circ$.\label{fig:slope_experiment}]{%
    \includegraphics[width=0.49\columnwidth]{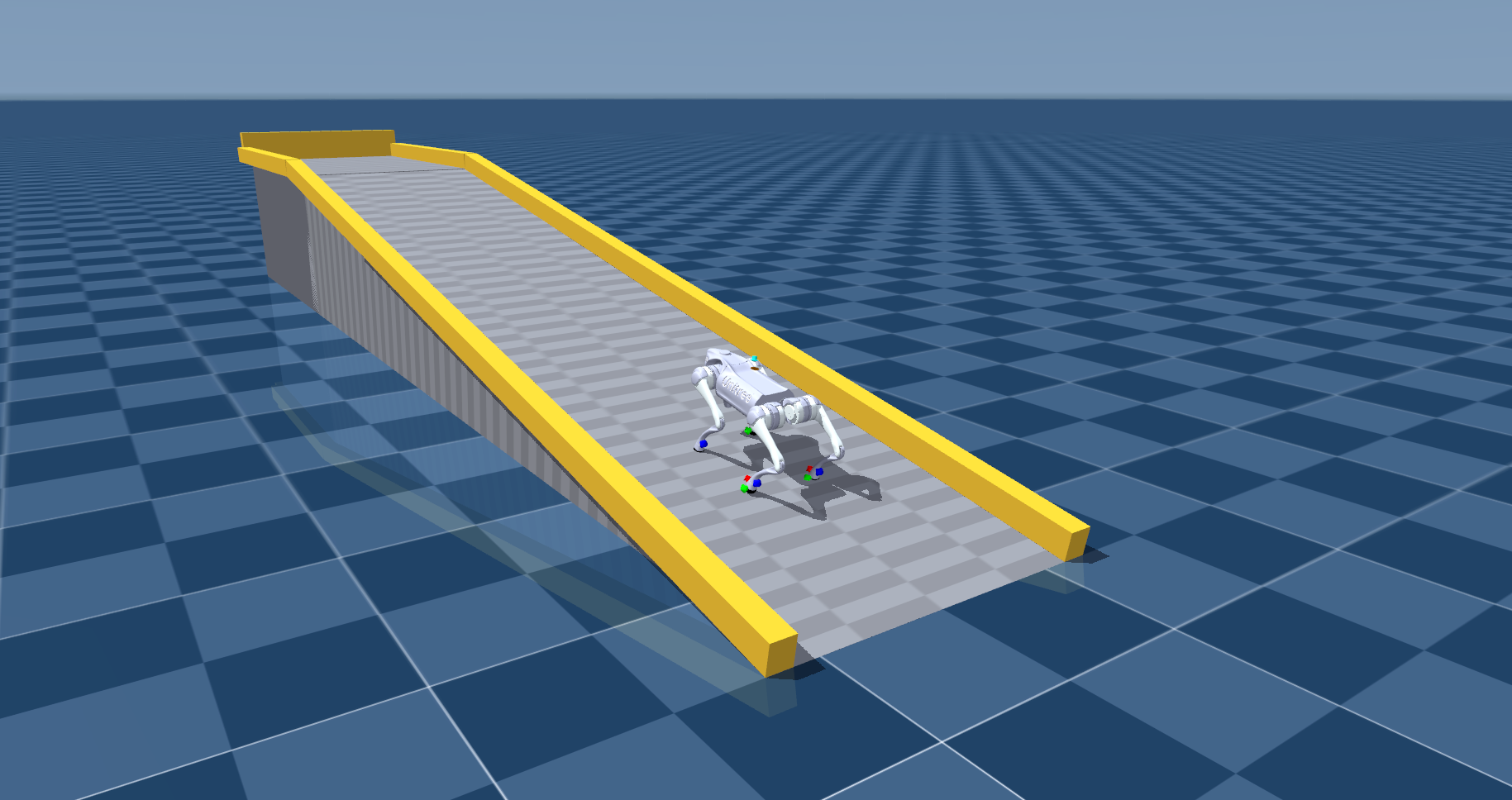}}
  \caption{Unitree Go2 simulation setups in MuJoCo.}
\end{figure}

\subsection{Simulated Experiments}
\label{ssec:sim_experiments}

The simulated evaluation involved both bipedal and quadrupedal robotic platforms,
specifically the Unitree \textit{G1} and \textit{Go2}.
These experiments were designed to evaluate the estimator under diverse locomotion
conditions and terrain geometries.

\subsubsection{Unitree Go2 MuJoCo Simulation, Rough Terrain}

To evaluate SBEE on quadrupedal platforms in challenging environments
and quantify the contribution
of the terrain-elevation update to the accuracy of the estimator,
we designed two simulation scenarios.
Particular emphasis was placed on the drift of the estimated base position
along the vertical axis. Locomotion commands for the Go2 were generated using
the adaptive nonlinear centroidal MPC proposed in~\cite{QuadrupedPYMPC2025}.

In the first scenario, the Go2 traversed rough terrain without access to exteroceptive measurements,
thereby operating under blind locomotion conditions.
The uneven terrain induces unmodeled events, including foot slippage and collisions between
the feet and terrain obstacles. The corresponding simulation setup is
shown in Fig.~\ref{fig:rough_experiment}.

\begin{figure}[t]
  \centering
  \includegraphics[width=0.8\columnwidth]
  {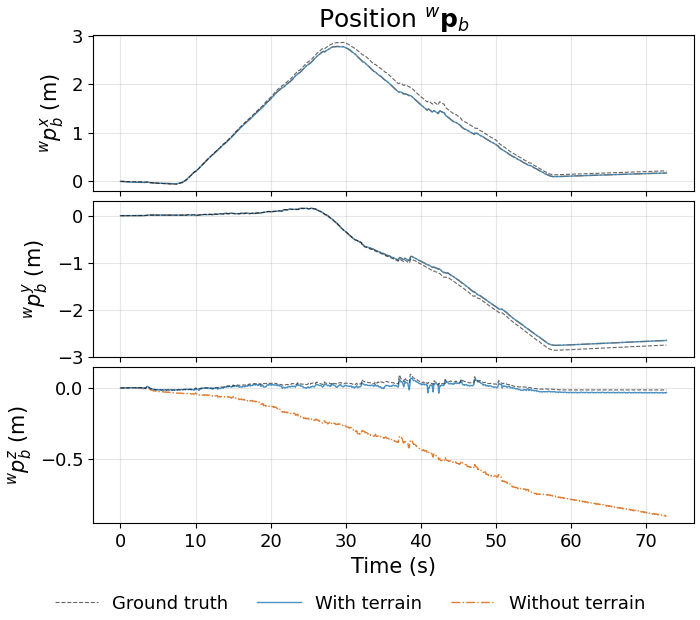}
  \caption{Base position estimates during the rough-terrain experiment.}
  \label{fig:pos_rough_experiment}
\end{figure}

\begin{figure}[b]
  \centering
  \includegraphics[width=0.8\columnwidth]
  {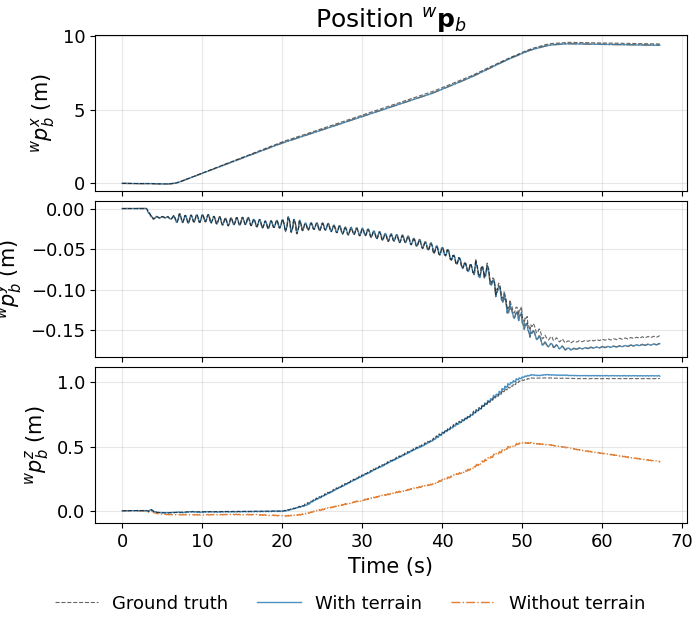}
  \caption{Base position estimates during the sloped-terrain experiment with Go2 robot.}
  \label{fig:slope_position}
\end{figure}

Fig.~\ref{fig:pos_rough_experiment} presents the qualitative comparison of
the estimated base position against the ground truth.
An ablation experiment was performed by comparing the complete SBEE framework, with the terrain-elevation
update enabled, against its underlying LI-EKF formulation with the terrain update disabled.

The effect of the terrain-elevation update is most evident in the vertical component of the base position.
As shown in Fig.~\ref{fig:pos_rough_experiment}, the complete SBEE framework prevents the vertical-position error from accumulating,
whereas the LI-EKF baseline accumulates substantial drift over time. The terrain-elevation measurement provides a
continuous correction to the vertical base position, which is unavailable when the terrain update is disabled.
Therefore, this ablation isolates the contribution of the terrain update within our formulation
and should not be read as indicative of the drift of contact-aided invariant filters that constrain the base relative to the stance foot.






\subsubsection{Unitree Go2 MuJoCo Simulation, Slope Terrain}
In the second scenario, Go2 traversed a sloped terrain with a $10^\circ$ incline.
The simulation setup is shown in Fig.~\ref{fig:slope_experiment}.
This scenario is complementary to the previous one, since the rough terrain experiment examines the ability of SBEE to bound drift over
a nominally flat surface while the slope introduces a genuine change in the elevation of the base, proving that the terrain elevation update
anchors the vertical axis without biasing the estimate toward a constant ground plane.
Fig.~\ref{fig:slope_position} presents the base position estimates. During the ascent, the base elevation from the ground is approximately $1.0\,$m.
The LI-EKF baseline recovers only about $0.35\,$m of this elevation gain, underestimating the final base height by
roughly $65\%$ ($0.65\,$m). The complete SBEE framework, in contrast, tracks the vertical position with a final error of $0.02\,$m.

\begin{figure}[t]
  \centering
  \includegraphics[width=0.9\columnwidth]
  {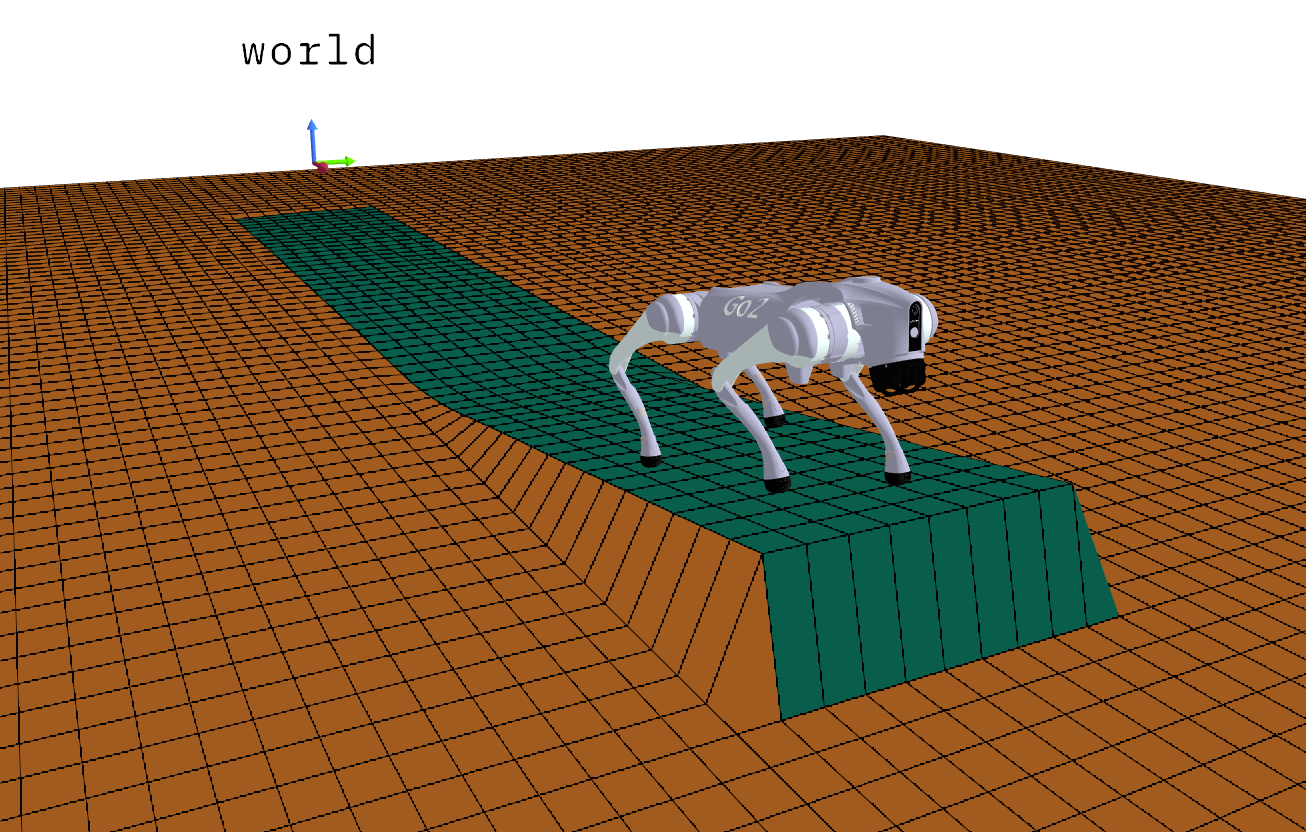}
  \caption{Estimated elevation map for the sloped-terrain simulation.}
  \label{fig:slope_elevation_map}
\end{figure}

The estimated elevation map accumulated during the traverse of the slope is depicted in Fig.~\ref{fig:slope_elevation_map}. The inclined surface is
recovered from the continuous grid map without any exteroceptive measurements and visualized using Foxglove~\cite{foxglove_studio}. In this scenario,
the orientation and linear velocity between the two configurations are indistinguishable, and therefore compared only quantitatively in Table ~\ref{tab:terrain_ablation}.

\begin{table*}[t]
  \centering
  \caption{Terrain-correction ablation on the Go2 and G1 simulations, metrics as in~\eqref{eq:vector_rmse},
    \eqref{eq:vertical_error} and~\eqref{eq:orientation_rmse}. Bold marks the better of each pair. The lower
    block is identical with and without the terrain update and is therefore reported once per experiment.}
  \label{tab:terrain_ablation}
  \setlength{\tabcolsep}{5pt}
  \renewcommand{\arraystretch}{1.15}
  \small
  \begin{tabular}{@{}lccccccccc@{}}
    \toprule
                                               & \multicolumn{3}{c}{Go2, slope} & \multicolumn{3}{c}{Go2, rough} & \multicolumn{3}{c}{G1, slope} \\
    \cmidrule(lr){2-4} \cmidrule(lr){5-7} \cmidrule(lr){8-10}
    Terrain Update                             & On & Off & Improvement & On & Off & Improvement & On & Off & Improvement \\
    \midrule
    \multicolumn{10}{@{}l}{\textit{Base position} ${}^{w}\bm{p}_{b}$} \\
    \quad $\operatorname{RMSE}_{3D}$ (m)       & \textbf{0.0810} & 0.3636  & 77.7\% & \textbf{0.1093}  & 0.5295  & 79.4\% & \textbf{0.0396} & 0.2226 & 82.2\% \\
    \quad $\operatorname{RMSE}_{z}$ (m)        & \textbf{0.0129} & 0.3547  & 96.4\% & \textbf{0.0168}  & 0.5184  & 96.8\% & \textbf{0.0253} & 0.2205 & 88.5\% \\
    \quad $\bar{e}_{z}$ (m)                    & \textbf{0.0046} & -0.2740 & 98.3\% & \textbf{-0.0144} & -0.4239 & 96.6\% & \textbf{0.0114} & 0.1769 & 93.6\% \\
    \addlinespace
    \multicolumn{10}{@{}l}{\textit{Unaffected by the terrain update}} \\
    \quad Position $\operatorname{RMSE}_{xy}$ (m)      & \multicolumn{3}{c}{0.0799} & \multicolumn{3}{c}{0.1080} & \multicolumn{3}{c}{0.0305} \\
    \quad Velocity $\operatorname{RMSE}_{3D}$ (m/s)    & \multicolumn{3}{c}{0.0225} & \multicolumn{3}{c}{0.0165} & \multicolumn{3}{c}{0.0640} \\
    \quad Velocity $\operatorname{RMSE}_{z}$ (m/s)     & \multicolumn{3}{c}{0.0162} & \multicolumn{3}{c}{0.0110} & \multicolumn{3}{c}{0.0186} \\
    \quad Velocity $\bar{e}_{z}$ (m/s)                 & \multicolumn{3}{c}{0.0092} & \multicolumn{3}{c}{0.0008} & \multicolumn{3}{c}{0.0132} \\
    \quad Orientation $\operatorname{RMSE}_{R}$ ($^\circ$) & \multicolumn{3}{c}{0.1313} & \multicolumn{3}{c}{0.5884} & \multicolumn{3}{c}{0.6576} \\
    \bottomrule
  \end{tabular}
\end{table*}

\begin{figure}[b]
  \centering
  \includegraphics[width=0.8\columnwidth]{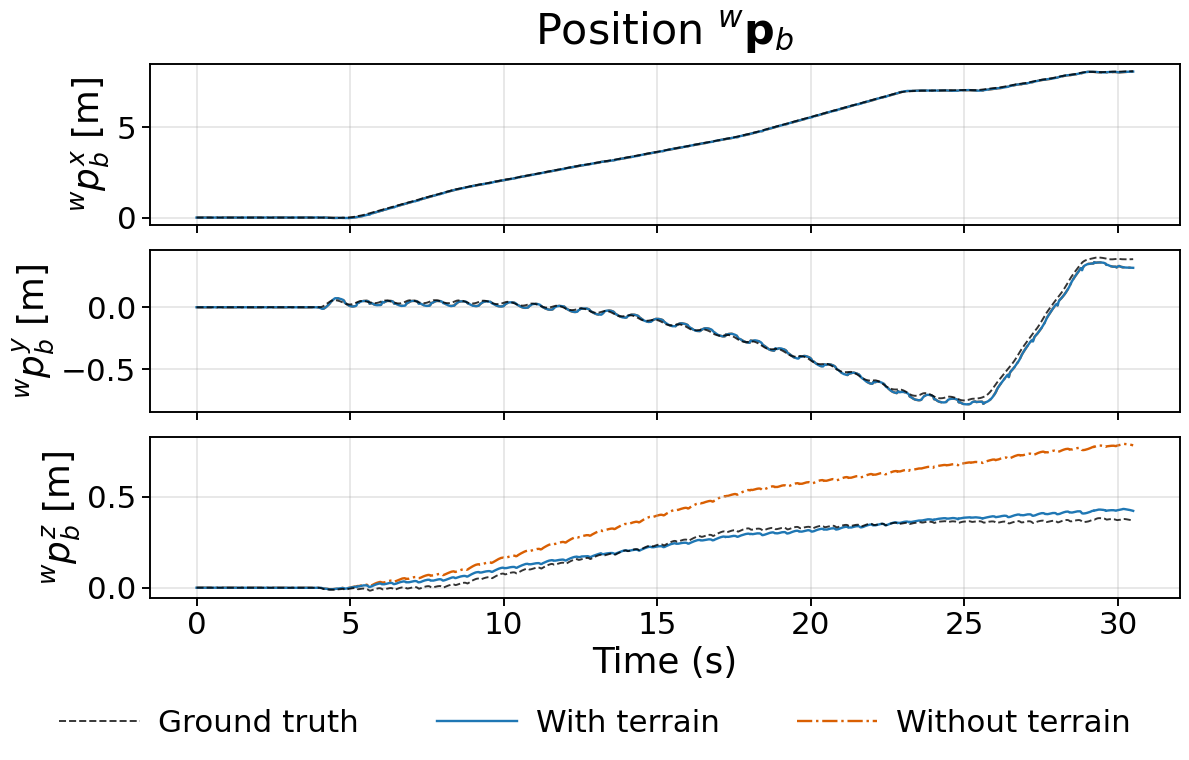}
  \caption{Base position estimates for the Unitree G1 during the sloped-terrain experiment.}
  \label{fig:g1_slope_position}
\end{figure}

The effect of the terrain-elevation update is largest along the vertical axis, on which it acts directly. The
vertical position error $\operatorname{RMSE}_{z}$ is reduced by $96.4\%$ on the Go2 slope and by $96.8\%$ on the rough
terrain, with comparable reductions of $98.3\%$ and $96.6\%$ in the accumulated offset $\bar{e}_{z}$. At the end
of the sequences, the residual vertical drift falls from $0.649$~m to $0.020$~m and from $0.887$~m to $0.019$~m,
respectively. The improvement in the full three-dimensional position RMSE is smaller, $77.7\%$ and $79.4\%$, since
it is dominated by the horizontal components.
The velocity and orientation estimates are identical in both configurations, since the partial terrain update
corrects the vertical base position only. The primary contribution of the terrain-elevation update is therefore
the suppression of accumulated vertical drift, with the remaining state estimates left untouched.
\begin{figure}[t]
  \centering
  \subfloat[Base orientation.]{
    \includegraphics[width=0.8\columnwidth]{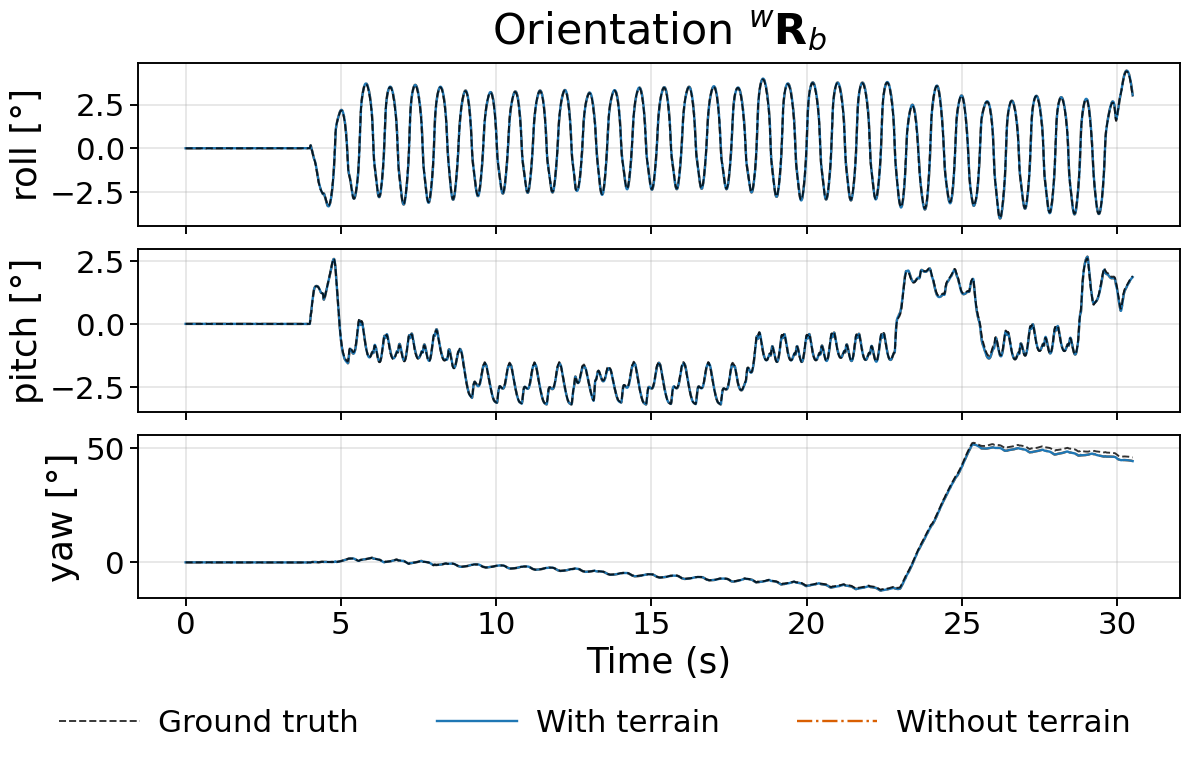}
    \label{fig:g1_slope_orientation}
  }
  \hfil
  \subfloat[Base linear velocity.]{
    \includegraphics[width=0.8\columnwidth]{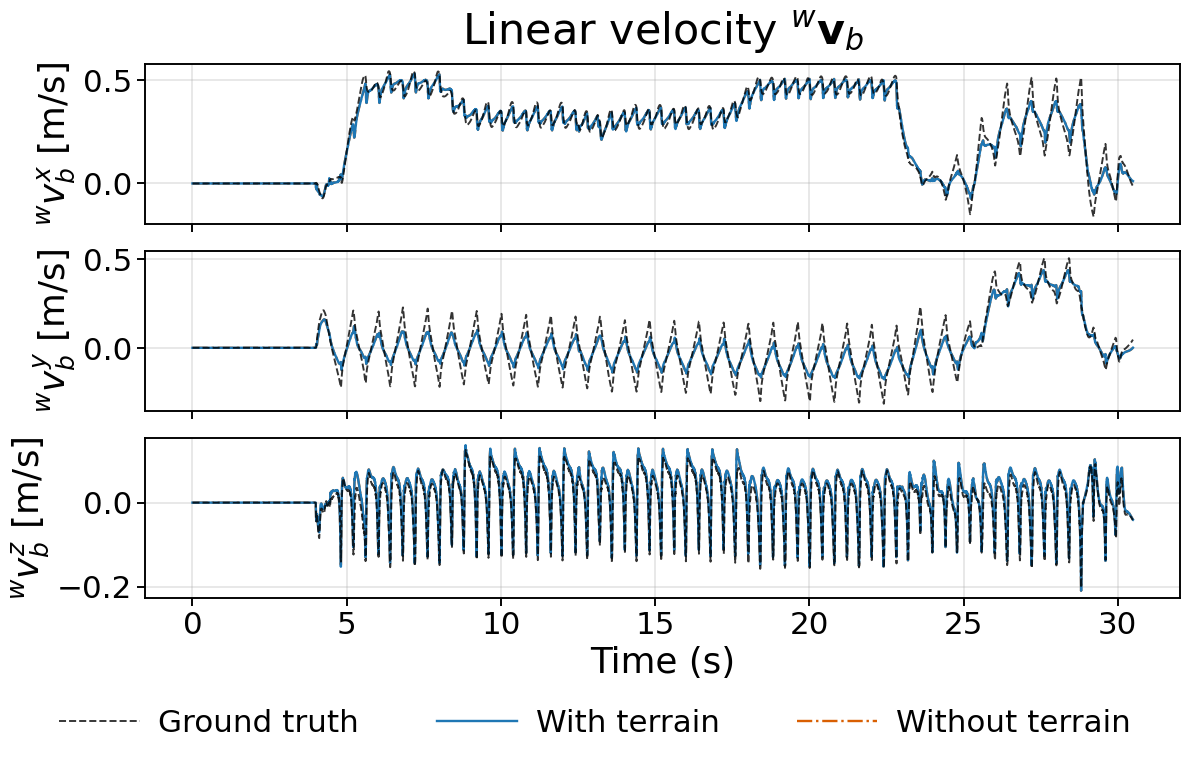}
    \label{fig:g1_slope_velocity}
  }
  \caption{Qualitative comparison against the ground truth during the G1 sloped-terrain experiment:
    (a) base orientation and (b) world frame base linear velocity.}
  \label{fig:g1_slope_qualitative}
\end{figure}

\subsubsection{Unitree G1 MuJoCo Simulation, Slope Terrain}
The third scenario evaluates SBEE on a bipedal platform with flat feet, namely Unitree's G1, traversing a
$10^\circ$ incline under the Unitree RL locomotion framework~\cite{unitree_rl_gym}. Only the terrain measurement differs, as each stable contact of the G1 anchors a
finite-area sole from which the surface normal is extracted and propagated over the local tangent plane
via~\eqref{eq:plane_proj}. This is the mechanism by which the map converges despite the substantially lower number
of contacts available on a biped, where at most one foot is in stable contact during single support.
Figure~\ref{fig:g1_slope_position} presents the base position estimates. The true base elevation rises by
approximately $0.37\,$m over the traverse, which the LI-EKF baseline overestimates with a mean vertical offset of
$+0.177\,$m, whereas the complete framework tracks the true profile to within $0.011\,$m. Notably, the sign of
the uncorrected drift is opposite to that observed on the quadrupedal platform, which confirms that the terrain
map anchors the vertical axis irrespective of the direction in which the drift accumulates. The horizontal
components are tracked closely by both configurations.
The orientation and world frame linear velocity are reported in Fig.~\ref{fig:g1_slope_qualitative}. Roll and
pitch exhibit the periodic oscillation characteristic of bipedal gait and the yaw follows the change of direction
executed near the end of the traverse, with the two configurations again visually indistinguishable. The estimated
lateral velocity ${}^{w}v_{b}^{y}$ shows a smaller oscillation amplitude than the ground truth, which is
attributed to the contact-confidence weighting of the kinematic measurement averaging across the double-support
transitions. The corresponding quantitative results are reported in Table~\ref{tab:terrain_ablation}.

\subsection{Comparative Evaluation on ANYmal-D GrandTour Dataset}
\label{ssec:anymal_grandtour}
To evaluate the proposed framework against representative
state-of-the-art proprioceptive estimators, we follow the recently published benchmarking
work in~\cite{nistico2026benchmark}. The experiments are
conducted on the CYN-1 sequence of the GrandTour Dataset
\cite{grandtour2026}, recorded with an ANYmal-D quadruped while traversing
the Grindelwald Canyon. The sequence contains over
$296\,$m of locomotion over uneven outdoor terrain and lasts
approximately $520\,$s.

SBEE is compared against the three estimators evaluated in
\cite{nistico2026benchmark}: MUSE in its proprioceptive-only
configuration~\cite{muse2025}, the contact-aided Invariant Extended Kalman Filter
(IEKF)~\cite{hartley2020contact}, and the fixed-lag Invariant Smoother (IS) with a window size of
three states~\cite{InvariantSmoother2024}. All methods use only proprioceptive measurements, namely
IMU measurements, joint encoders, leg kinematics, and contact
information. More specifically, the dataset provides binary contact flags for
each foot that were utilized by the aforementioned estimators. However, SBEE requires a
continuous contact confidence to weight the kinematic velocity and terrain elevation updates. Therefore,
we have utilized the joint efforts alongside with a Generalized Momentum Observer (GMO)~\cite{DeLuca2006} to estimate the contact wrench
and utilized the estimated forces to compute the contact confidence as described in Section~\ref{sec:contact_est}. The contact confidence is then used to weight the kinematic velocity and terrain elevation updates.
The published benchmark implementations and preprocessing
pipeline were used to reproduce the MUSE, IEKF, and IS results on the
same sequence.

The published ATE and RPE results for MUSE, IEKF, and IS were
reproduced exactly using the released benchmarking pipeline. During
the reproduction of the linear-velocity metric, however, we identified
a timestamp-association issue in the publicly released evaluation
script. The issue was also reported through the public repository of
the original benchmark.

The estimated and ground truth trajectories are evaluated over their
common temporal interval. Following~\cite{nistico2026benchmark}, we
report the following metrics:

\begin{itemize}
  \item \textbf{Absolute Trajectory Error (ATE):}
        the position RMSE after rigid $\SOthree$ Umeyama alignment of the
        estimated trajectory to the ground truth. ATE measures long-term
        trajectory consistency and accumulated drift.

  \item \textbf{Linear-velocity RMSE:}
        the RMSE between synchronized estimated and ground truth
        world frame velocity samples. The rotational component of the
        trajectory alignment is applied to the estimated velocity before
        comparison.

  \item \textbf{Relative Pose Error (RPE):}
        translational and rotational errors evaluated using two different
        intervals. A spatial interval of $\Delta=1\,\mathrm{m}$ measures
        local drift accumulated over meter-scale trajectory segments,
        whereas $\Delta=1$ frame measures frame-to-frame consistency and
        sensitivity to high-frequency measurement fluctuations.
\end{itemize}

ATE and RPE are computed using the \texttt{evo} evaluation
package~\cite{evo2017}, using the same trajectory conversion,
interpolation, and alignment procedure as the reference benchmark.

\begin{figure*}[!t]
  \centering
  \includegraphics[width=\textwidth]{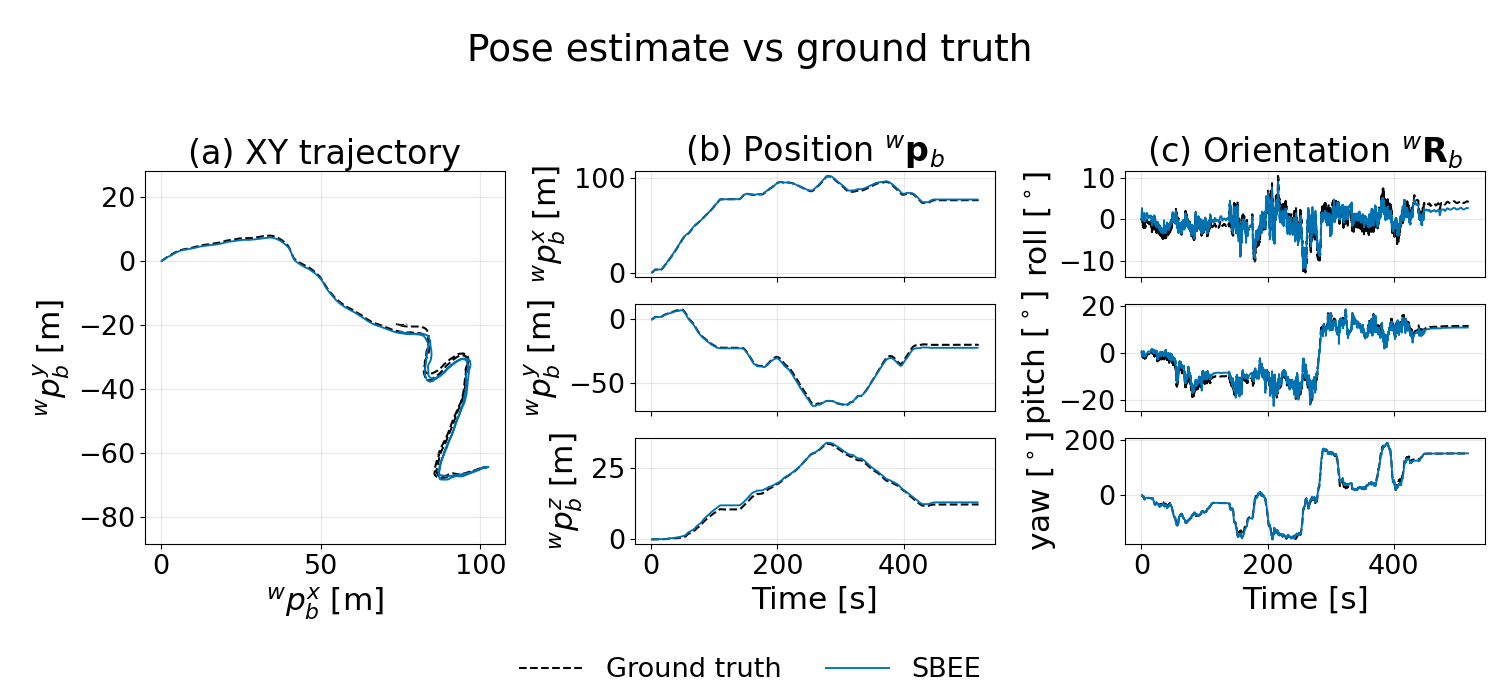}
  \caption{Comparison of the SBEE pose estimate and ground truth on the
    ANYmal-D CYN-1 sequence. Left: horizontal trajectory. Middle:
    world frame position components. Right: roll, pitch, and yaw angles.}
  \label{fig:anymal_pose}
\end{figure*}

\subsubsection{Quantitative Results}

Table~\ref{tab:anymal_benchmark} reports the quantitative results. The
best result in each row is shown in bold, while the second-best result
is underlined.
\begin{table}[!b]
  \centering
  \caption{Comparative evaluation between SBEE and three state-of-the-art proprioceptive estimators on the ANYmal-D GrandTour Dataset.}
  \label{tab:anymal_benchmark}
  \setlength{\tabcolsep}{3.2pt}
  \renewcommand{\arraystretch}{1.12}
  \begin{tabular}{lcccc}
    \toprule
    \textbf{RMSE metric}
     & \textbf{MUSE~\cite{muse2025}}
     & \textbf{IEKF~\cite{hartley2020contact}}
     & \textbf{IS~\cite{InvariantSmoother2024}}
     & \textbf{SBEE}                            \\
    \midrule
    ATE [m]
     & 2.269461
     & 1.405668
     & \underline{1.363114}
     & \textbf{1.062045}                        \\

    Velocity RMSE [m/s]
     & 0.085253
     & 0.081476
     & \underline{0.081454}
     & \textbf{0.075819}                        \\

    RPE ($\Delta=1\,\mathrm{m}$) [m]
     & 0.072223
     & \underline{0.043187}
     & \textbf{0.042526}
     & 0.051362                                 \\

    RPE ($\Delta=1$ frame) [m]
     & \underline{0.000670}
     & \textbf{0.000544}
     & 0.001965
     & 0.001071                                 \\

    RPE ($\Delta=1\,\mathrm{m}$) [$^{\circ}$]
     & 0.578469
     & 0.568987
     & \underline{0.565644}
     & \textbf{0.556790}                        \\

    RPE ($\Delta=1$ frame) [$^{\circ}$]
     & \textbf{0.002605}
     & \underline{0.002611}
     & 0.002614
     & 0.003722                                 \\
    \bottomrule
  \end{tabular}

  \vspace{1mm}
\end{table}

SBEE achieves the lowest ATE, with an error of
$1.062\,\mathrm{m}$. This corresponds to a reduction of approximately
$22.1\%$ relative to the strongest baseline, the IS, and reductions of
$24.4\%$ and $53.2\%$ relative to the IEKF and MUSE, respectively.
Considering that the evaluated trajectory is over
$296\,\mathrm{m}$ long, the SBEE ATE corresponds to approximately
$0.36\%$ of the total travelled distance. The result indicates that the
proposed estimator accumulates less long-term trajectory drift than the
three reference estimators.

SBEE also obtains the lowest linear-velocity RMSE,
$0.0758\,\mathrm{m/s}$, representing an improvement of approximately
$6.9\%$ over the IS and IEKF results and $11.1\%$ over MUSE. This
indicates that the fixed-size LI-EKF and its contact-confidence-weighted
kinematic velocity updates accurately reproduce the global velocity
profile throughout the sequence.

The relative-pose results reveal a more nuanced trade-off. For
translational RPE at $\Delta=1\,\mathrm{m}$, SBEE achieves
$0.0514\,\mathrm{m}$. This is $28.9\%$ lower than MUSE, but remains
higher than the IEKF and IS by approximately $18.9\%$ and $20.8\%$,
respectively. Thus, although SBEE provides the best full-trajectory
accuracy, the contact-state IEKF and fixed-lag smoother retain
better translational consistency over one-meter segments.

At the frame-to-frame scale, SBEE obtains a translational RPE of
$0.001071\,\mathrm{m}$. This is lower than the IS result but higher than
MUSE and the IEKF. The small increase relative to the two filtering
baselines suggests that the stronger velocity and terrain corrections
may introduce additional high-frequency variation while reducing
low-frequency accumulated drift. This interpretation is consistent with
the difference between the long-horizon ATE and short-horizon RPE
metrics: ATE rewards long-term consistency, whereas one-frame RPE is
particularly sensitive to estimator bandwidth, contact transitions, and
measurement jitter.

For rotational RPE at $\Delta=1\,\mathrm{m}$, SBEE achieves the best
result, $0.5568^{\circ}$, improving upon the IS, IEKF, and MUSE by
approximately $1.6\%$, $2.1\%$, and $3.7\%$, respectively. Conversely,
its frame-to-frame rotational RPE is $0.003722^{\circ}$, approximately
$43\%$ higher than those of the three reference estimators. The absolute
magnitude nevertheless remains very small. The result again indicates
that SBEE preserves accurate orientation over longer trajectory
segments while exhibiting slightly greater high-frequency variation.

Overall, the quantitative comparison shows that SBEE provides the best
long-term position accuracy, velocity accuracy, and meter-scale
rotational consistency. The gains are not uniform across all metrics:
the IEKF and IS remain superior in meter-scale translational RPE, while
MUSE and the IEKF achieve lower frame-to-frame errors. This demonstrates
that the principal advantage of SBEE lies in suppressing accumulated
drift rather than exclusively minimizing instantaneous motion
increments.

\subsubsection{Trajectory and Orientation Estimates}

Figure~\ref{fig:anymal_pose} compares the SBEE pose estimate with the
ground truth. The estimated horizontal trajectory remains closely aligned
with the reference over the complete sequence, including the long initial
traverse, the curved canyon section, and the repeated changes of
direction near the end of the trajectory. The remaining horizontal
deviation is smooth and limited, consistent with the low ATE reported in
Table~\ref{tab:anymal_benchmark}.
The per-axis position plots further show that SBEE follows the large
variations in ${}^{w}p_{b}^x$ and ${}^{w}p_{b}^y$ without discontinuities. Of particular
interest is the vertical position ${}^{w}p_{b}^z$, which changes by more than
$30\,\mathrm{m}$ as the robot traverses the canyon. SBEE closely follows
the complete ascent and subsequent descent and terminates close to the
ground truth elevation. This result demonstrates that the terrain-assisted
corrections do not force the estimate toward a constant ground plane and
remain compatible with substantial genuine changes in terrain elevation.

The estimated roll, pitch, and yaw also follow the ground truth over the
entire experiment. The largest roll and pitch discrepancies occur during
the most dynamically excited sections of the sequence, but the estimates
remain bounded and rapidly return toward the reference. The yaw profile
closely follows the ground truth relative rotation after trajectory
alignment. This should not be interpreted as absolute yaw observability:
global yaw remains unobservable from proprioceptive measurements alone,
and its absolute offset is removed by the trajectory alignment.


\subsubsection{Linear-Velocity Estimates}

\begin{figure}[!t]
  \centering
  \includegraphics[width=\columnwidth]{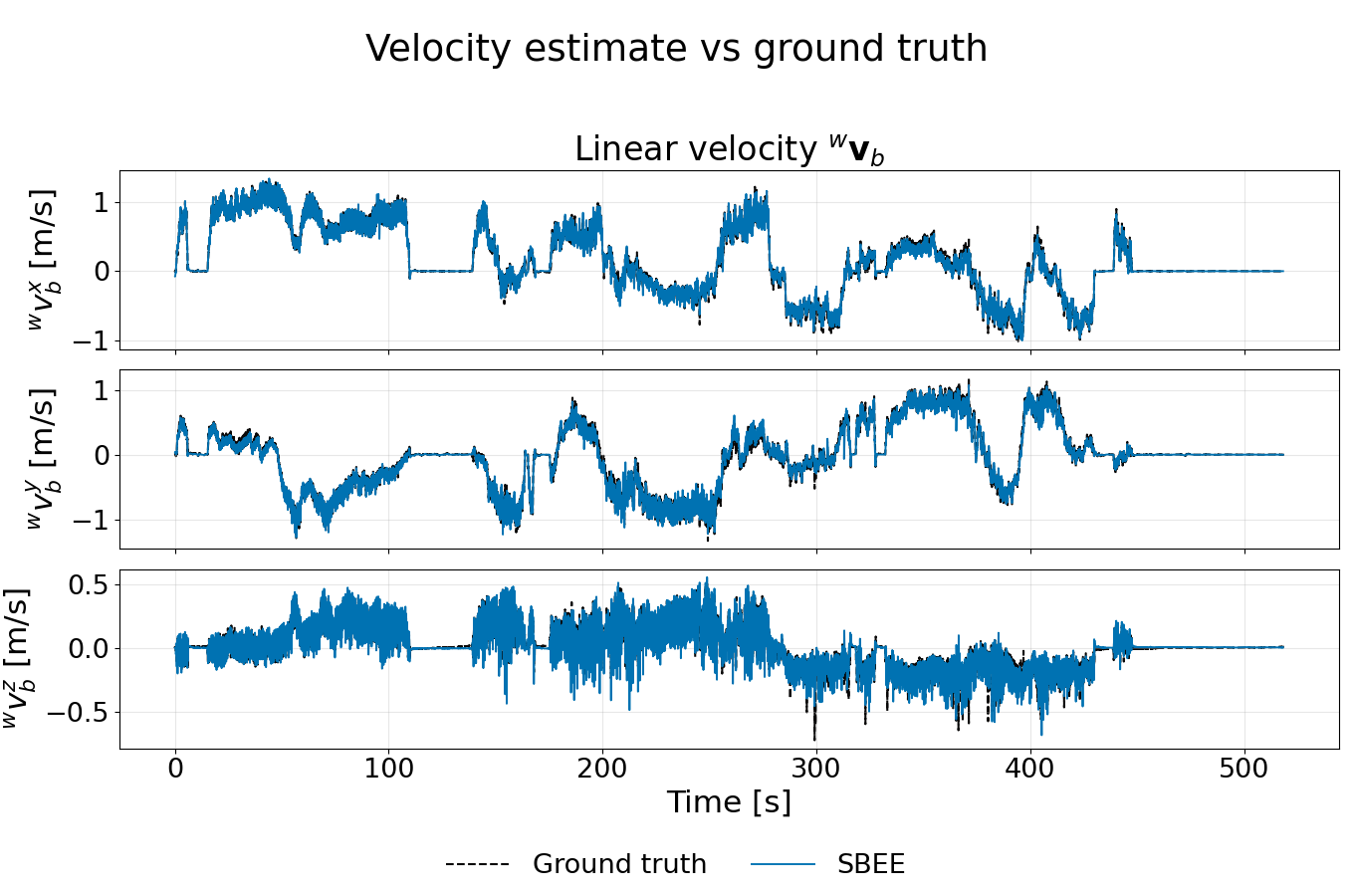}
  \caption{Comparison of the SBEE world frame linear-velocity estimate
    and ground truth on the ANYmal-D CYN-1 sequence. From top to bottom:
    ${}^{w}v_{b}^x$, ${}^{w}v_{b}^y$, and ${}^{w}v_{b}^z$.}
  \label{fig:anymal_velocity}
\end{figure}

The estimated world frame velocity components are presented in
Fig.~\ref{fig:anymal_velocity}. SBEE closely follows the ground truth
${}^{w}v_{b}^x$ and ${}^{w}v_{b}^y$ profiles throughout the sequence, including rapid
transitions between stationary and locomotion phases and repeated
changes in direction. This close agreement is reflected in the lowest
velocity RMSE among the evaluated estimators.

The vertical velocity ${}^{w}v_{b}^z$ exhibits greater high-frequency variation
than the horizontal components. This is expected because vertical
velocity is particularly sensitive to accelerometer noise, attitude
uncertainty, contact transitions, and errors in the reconstructed
foot velocity. Nevertheless, its mean evolution and major dynamic
variations remain consistent with the ground truth, and no persistent
velocity offset is observed. Preventing such an offset is important
because even a small vertical-velocity bias would integrate into
substantial base-height drift over the duration of the experiment.

\subsubsection{Contact-Confidence Estimation}

\begin{figure}[!t]
  \centering
  \includegraphics[width=0.95\columnwidth]{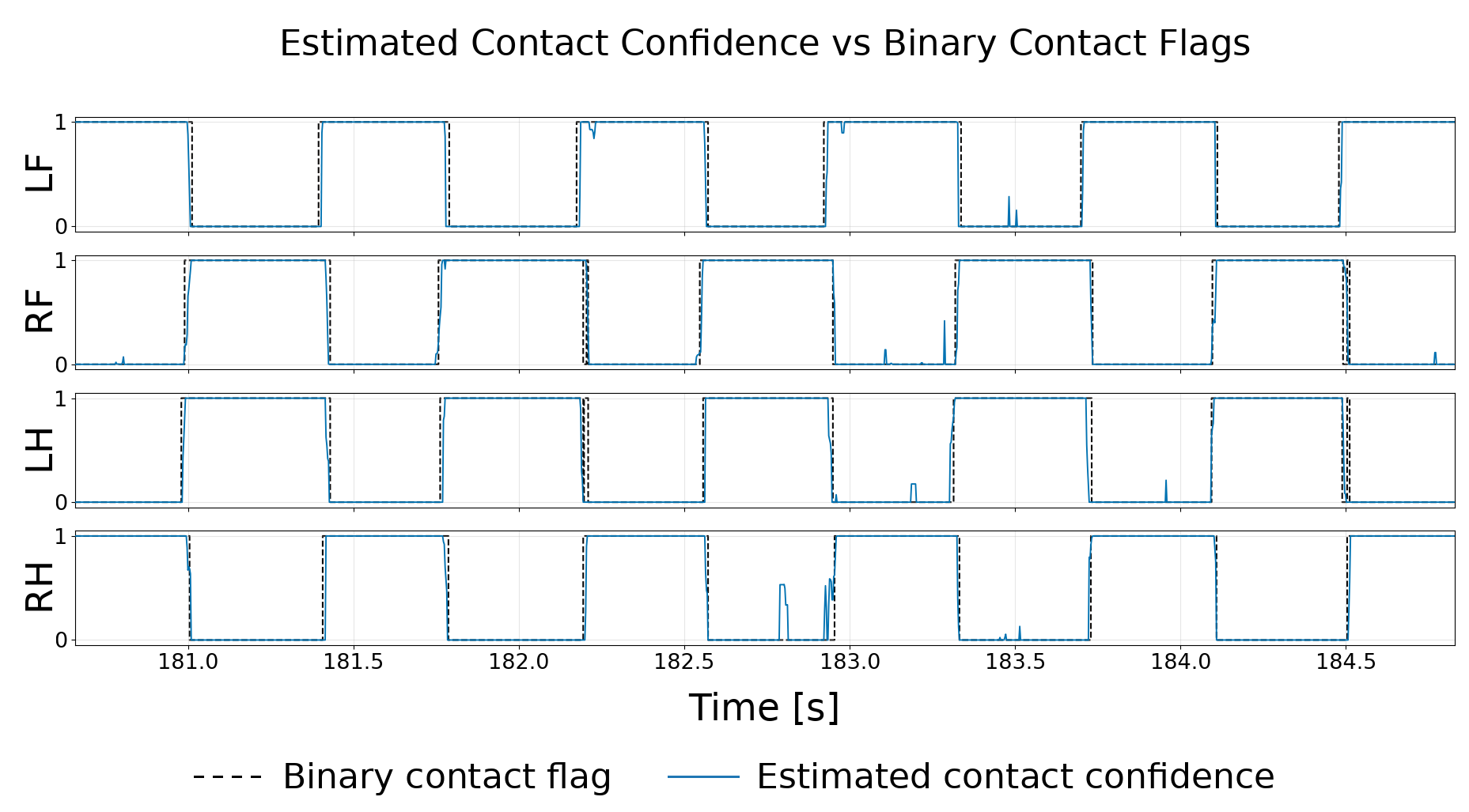}
  \caption{SBEE-estimated contact confidence compared with the
    reference binary contact flags for the left-front (LF), right-front
    (RF), left-hind (LH), and right-hind (RH) feet over a representative
    interval of the CYN-1 sequence.}
  \label{fig:anymal_contacts}
\end{figure}

Figure~\ref{fig:anymal_contacts} compares the continuous contact
confidence values used by SBEE against the binary contact flags available
for the CYN-1 sequence over a representative locomotion interval. The
estimated contact state closely reproduces the reference stance and
swing phases for all four feet.

Short fractional deviations occur primarily around contact transitions
and during isolated dynamic events. These deviations are expected
because SBEE intentionally represents contact as a continuous confidence
rather than imposing an instantaneous binary switch. The probabilistic
representation allows uncertain contacts to contribute proportionally
to the kinematic velocity and terrain updates, reducing abrupt estimator
changes caused by hard contact switching. At the same time, the observed
short transients may partly explain the slightly higher one-frame RPE,
since rapid changes in contact confidence directly affect the effective
velocity measurement and its covariance.

The comparison evaluates the complete SBEE pipeline and therefore does
not, by itself, isolate the individual contributions of the
left-invariant formulation and the terrain-elevation feedback. These
effects are examined separately through the ablation experiments
presented in Section~\ref{ssec:sim_experiments}.

\subsection{Computational Performance}
The computational cost of SBEE was measured on two platforms: a desktop workstation with a 12th-generation Intel
Core i7-12700K and $32\,$GB of RAM, and a Raspberry~Pi~5 with $8\,$GB of RAM, representative of the embedded
computers with which most legged robots are equipped. The reported results of Table~\ref{tab:timing} are produced from offline data executed in a
single thread with terrain elevation update enabled and disabled to extract the impact it has on each cycle update time.

Table~\ref{tab:timing} summarizes the results. On both platforms the estimator's refresh rate is well beyond the refresh rate required by
high-rate locomotion controller. Enabling the terrain elevation update adds between $4\%$ and $14\%$ of the cycle time, which is attributed to the
scalar covariance correction rather than the map itself whose constant time toroidal indexing (Section~\ref{ssec:toroidal}) and scalar cell updates
are negligible. For reference,~\cite{nistico2026benchmark} reports per-iteration runtimes of $0.012\,$ms for MUSE, $0.020\,$ms for the IEKF
and $0.110$--$0.583\,$ms for the invariant smoother. A direct comparison is not possible because the hardware utilized to produce these results
is unspecified and our cycle additionally covers kinematics, contact estimation and leg odometry.

\begin{table}[b]
  \centering
  \caption{Computational performance of SBEE}
  \label{tab:timing}
  \begin{tabular}{llccc}
    \toprule
    Platform       & Terrain & Cycle  & Refresh rate \\
                   &         & (ms)   & (kHz)        \\
    \midrule
    Core i7-12700K & on      & 0.0614 & 16.3         \\
                   & off     & 0.0540 & 18.5         \\
    \addlinespace
    Raspberry Pi 5 & on      & 0.1617 & 6.2          \\
                   & off     & 0.1549 & 6.5          \\
    \bottomrule
  \end{tabular}
\end{table}

\section{Discussion}
\label{sec:discussion}
The advantages of SBEE follow from the interaction of its two components rather than from either in isolation:

\begin{enumerate}
  \item \textbf{State-independent linearization.}
Because the process dynamics are group-affine, the propagation Jacobian~\eqref{eq:AX_left} is independent of
the estimated state, so estimation errors corrupt neither the prediction nor the velocity linearization, which
keeps the propagated vertical covariance consistent. This matters for the coupling, since that covariance is
precisely what the terrain gain is weighed against.
\item \textbf{Scalar, partial terrain coupling.}
      Because the left-invariant error expresses the position error in the local base frame, the world frame
      height residual maps onto it through a single row, as in~\eqref{eq:H_terrain}, and the Schmidt--Kalman gain~\eqref{eq:K_terrain} confines the correction to that row. The map elevation therefore enters as
      a scalar update that moves the base vertically and nothing else: horizontal position, attitude and IMU
      biases are provably untouched (Table~\ref{tab:terrain_ablation}), no matrix inversion is required, and
      an erroneous map cell can at worst bias the height, never destabilise the attitude or bias estimates.
  \item \textbf{Bounded computational cost.}
        The map is stored as a fixed-size array addressed through toroidal indexing, so that the modulo reduces to
        a bitwise AND and recentering shifts only the map origin, leaving accumulated elevation data untouched in
        memory. Together with the fixed state dimension, this keeps the cost of a cycle constant in the number of
        legs, and confines it to a fraction of the control budget even on embedded hardware
        (Table~\ref{tab:timing}).
\end{enumerate}
A few aspects remain open for future work.
The framework assumes static, non-deformable terrain: soft ground or obstacles displaced after being mapped inject
an erroneous height reference, whose influence persists until repeated contacts revise the affected cells. The
terrain feedback bounds the vertical axis only, as the horizontal position and the yaw angle remain unobservable
from proprioception alone. The contact confidence of~\eqref{eq:contact_prob} is derived from the vertical ground reaction force and
therefore cannot detect slippage, although the modular design of the estimator allows any continuous contact
estimator to be used in its place.

\section{Conclusions}
\label{sec:conclusions}

We have presented Simultaneous Base and Elevation Estimation (SBEE), a proprioceptive framework in which a legged robot
builds a tactile elevation map of its environment and uses
that map to anchor its vertical position estimate. The terrain elevation mapper maintains a probabilistic grid that is
efficiently shifted through toroidal indexing, and contact observations are fused into it by scalar Kalman
filtering with planar neighborhood propagation. A Left-Invariant EKF on $\SEtwo \times \R^{6}$ queries this map to
obtain pseudo-absolute altitude measurements whose strength depends on how well the terrain has been mapped: fresh
terrain yields gentle corrections, whereas returning to well-mapped terrain produces sharp, loop-closure-like
corrections that eliminate accumulated vertical drift. Because the contact memory resides in the map rather than in
augmented contact states, the estimator retains a fixed state dimension and can be deployed unchanged to platforms with an
arbitrary number of legs and to both point-foot and flat-foot morphologies.

On the ANYmal-D GrandTour sequence, SBEE attains an absolute trajectory error of $1.062\,$m over $296\,$m of
locomotion, improving upon the strongest proprioceptive baseline by $22.1\%$, at a cost of $0.162\,$ms per
estimation cycle on a Raspberry~Pi~5. Vertical position, otherwise unobservable from proprioception, is thus
rendered bounded and accurate without any exteroceptive sensor. Our implementation will be released as
open-source upon acceptance.

\appendices
\section{Derivation of the Left-Invariant Error Dynamics}
\label{sec:appendix}

This appendix details the derivation of the linearized error dynamics
in~\eqref{eq:log_linear_property}--\eqref{eq:AX_left}, obtained by
expanding the autonomous left-invariant error~\eqref{eq:autonomous_left_error} to first-order in
$\bm{\xi}_t$.

Recall the left-invariant error and its first-order approximation are:
\begin{equation}
  \bm{\eta}_{t}^{L} = \mathcal{X}_{t}^{-1}\hat{\mathcal{X}}_{t}= \Exp\!\bigl(\bm{\xi}_{t}\bigr)
  \approx
  \bm{I} + \bm{\xi}_{t}^{\wedge}
  \label{eq:app_eta_approx}
\end{equation}
From~\eqref{eq:autonomous_left_error}--\eqref{eq:left_error_function}:
\begin{equation}
  \frac{d}{dt}\bm{\eta}_{t}^{L}
  =
  g_{\bm{u}_{t}}^{L}\bigl(\bm{\eta}_{t}^{L}\bigr)
  +
  \bm{w}_{t}^{\wedge}\bm{\eta}_{t}^{L}
  \label{eq:app_eta_dot}
\end{equation}
with
$g_{\bm{u}_{t}}^{L}(\bm{\eta}_{t}^{L})
  =
  f_{\bm{u}_{t}}(\bm{\eta}_{t}^{L})
  -
  f_{\bm{u}_{t}}(\bm{I})\,\bm{\eta}_{t}^{L}$.

Substituting~\eqref{eq:app_eta_approx}
into~\eqref{eq:app_eta_dot} and dropping the second-order terms gives:
\begin{align}
  \frac{d}{dt}\bm{\xi}_{t}^{\wedge}
   & =
  g_{\bm{u}_{t}}^{L}\bigl(\bm{I}+\bm{\xi}_{t}^{\wedge}\bigr)
  +
  \bm{w}_{t}^{\wedge}
  \bigl(\bm{I}+\bm{\xi}_{t}^{\wedge}\bigr)
  \nonumber \\
   & =
  f_{\bm{u}_{t}}\bigl(\bm{I}+\bm{\xi}_{t}^{\wedge}\bigr)
  -
  f_{\bm{u}_{t}}(\bm{I})
  \bigl[\bm{I}+\bm{\xi}_{t}^{\wedge}\bigr]
  +
  \bm{w}_{t}^{\wedge}
  +
  \bm{w}_{t}^{\wedge}\bm{\xi}_{t}^{\wedge}
  \nonumber \\
   & \simeq
  f_{\bm{u}_{t}}\bigl(\bm{I}+\bm{\xi}_{t}^{\wedge}\bigr)
  -
  f_{\bm{u}_{t}}(\bm{I})
  -
  f_{\bm{u}_{t}}(\bm{I})\,\bm{\xi}_{t}^{\wedge}
  +
  \bm{w}_{t}^{\wedge}
  \label{eq:app_master}
\end{align}
where the last line neglects the second-order noise-error coupling
term $\bm{w}_{t}^{\wedge}\bm{\xi}_{t}^{\wedge}$.
The perturbed group element is:
\begin{equation}
  \bm{I}+\bm{\xi}_{t}^{\wedge}
  =
  \begin{pmatrix}
    \bI_{3}+\skewmat{\bm{\xi}_{t}^{R}}
                      & \bm{\xi}_{t}^{v} & \bm{\xi}_{t}^{p} \\
    \bm{0}_{1\times3} & 1                & 0                \\
    \bm{0}_{1\times3} & 0                & 1
  \end{pmatrix}
  \label{eq:app_perturbed_group}
\end{equation}
and, from the definition of $f_{\bm{u}_{t}}(\cdot)$
in~\eqref{eq:group_dynamics_compact}, the three terms
of~\eqref{eq:app_master} evaluate to:
\small
\begin{align}
  f_{\bm{u}_{t}}\bigl(\bm{I}+\bm{\xi}_{t}^{\wedge}\bigr)
   & =
  \begin{pmatrix}
    \skewmat{\bm{\omega}_{t}}+\skewmat{\bm{\xi}_{t}^{R}}\skewmat{\bm{\omega}_{t}}
                      &
    \bm{a}_{t}+\bg+\skewmat{\bm{\xi}_{t}^{R}}\bm{a}_{t}
                      &
    \bm{\xi}_{t}^{v}
    \\
    \bm{0}_{1\times3} & 0 & 0 \\
    \bm{0}_{1\times3} & 0 & 0
  \end{pmatrix}
  \label{eq:app_f_perturbed}
  \\[4pt]
  f_{\bm{u}_{t}}(\bm{I})
   & =
  \begin{pmatrix}
    \skewmat{\bm{\omega}_{t}}
                      & \bm{a}_{t}+\bg
                      & \bm{0}_{3\times1}
    \\
    \bm{0}_{1\times3} & 0                 & 0 \\
    \bm{0}_{1\times3} & 0                 & 0
  \end{pmatrix}
  \label{eq:app_f_identity}
  \\[4pt]
  f_{\bm{u}_{t}}(\bm{I})\,\bm{\xi}_{t}^{\wedge}
   & =
  \begin{pmatrix}
    \skewmat{\bm{\omega}_{t}}\skewmat{\bm{\xi}_{t}^{R}}
                      &
    \skewmat{\bm{\omega}_{t}}\bm{\xi}_{t}^{v}
                      &
    \skewmat{\bm{\omega}_{t}}\bm{\xi}_{t}^{p}
    \\
    \bm{0}_{1\times3} & 0 & 0 \\
    \bm{0}_{1\times3} & 0 & 0
  \end{pmatrix}
  \label{eq:app_f_identity_xi}
\end{align}
\normalsize
Substituting~\eqref{eq:app_f_perturbed}--\eqref{eq:app_f_identity_xi}
into~\eqref{eq:app_master}, the remainder collects as:
\begin{equation}
  \begin{split}
    \frac{d}{dt}\bm{\xi}_{t}^{\wedge}
    = {}        & \begin{pmatrix}
                    \bm{A}_{11}       & \bm{A}_{12} & \bm{A}_{13} \\
                    \bm{0}_{1\times3} & 0           & 0           \\
                    \bm{0}_{1\times3} & 0           & 0
                  \end{pmatrix}
    + \bm{w}_{t}^{\wedge}                                                                                                     \\[0.5em]
    \text{with}\quad
    \bm{A}_{11} & = \skewmat{\bm{\xi}_{t}^{R}}\skewmat{\bm{\omega}_{t}} - \skewmat{\bm{\omega}_{t}}\skewmat{\bm{\xi}_{t}^{R}} \\
    \bm{A}_{12} & = \skewmat{\bm{\xi}_{t}^{R}}\bm{a}_{t} - \skewmat{\bm{\omega}_{t}}\bm{\xi}_{t}^{v}                          \\
    \bm{A}_{13} & = \bm{\xi}_{t}^{v} - \skewmat{\bm{\omega}_{t}}\bm{\xi}_{t}^{p}
  \end{split}
  \label{eq:app_collected}
\end{equation}
Using the $\SO$ Lie-bracket identity
$\skewmat{\bm{\xi}_{t}^{R}}\skewmat{\bm{\omega}_{t}}
  -
  \skewmat{\bm{\omega}_{t}}\skewmat{\bm{\xi}_{t}^{R}}
  =
  -\skewmat{\skewmat{\bm{\omega}_{t}}\bm{\xi}_{t}^{R}}$,
Eq.~\eqref{eq:app_collected} can equivalently be written as a single
wedge map applied to $\bm{\xi}_{t}$:
\begin{equation}
  \frac{d}{dt}\bm{\xi}_{t}^{\wedge}
  =
  \left[
    \begin{pmatrix}
      -\skewmat{\bm{\omega}_{t}} & \bm{0}_{3}                 & \bm{0}_{3}                 \\
      -\skewmat{\bm{a}_{t}}      & -\skewmat{\bm{\omega}_{t}} & \bm{0}_{3}                 \\
      \bm{0}_{3}                 & \bI_{3}                    & -\skewmat{\bm{\omega}_{t}}
    \end{pmatrix}
    \bm{\xi}_{t}
    \right]^{\wedge}
  +
  \bm{w}_{t}^{\wedge}
  \label{eq:app_wedge_form}
\end{equation}
Applying the vee operator $(\cdot)^{\vee}$
to~\eqref{eq:app_wedge_form} and substituting IMU
measurements $\bm{\omega}_{t}\!\to\!\bm{\omega}_{t}^{\mathrm{imu}}$
and $\bm{a}_{t}\!\to\!\bm{a}_{t}^{\mathrm{imu}}$ recovers the linear
error dynamics of~\eqref{eq:log_linear_property}:
\begin{equation}
  \frac{d}{dt}\bm{\xi}_{t} = \bA_{t}\,\bm{\xi}_{t} + \bm{w}_{t}
  \label{eq:app_final_linear}
\end{equation}
with the state Jacobian $\bA_{t}$ exactly as given
in~\eqref{eq:AX_left}. This confirms that, to first order, the
left-invariant error dynamics are independent of the true and
estimated trajectories and depend only on the IMU input
$\bm{u}_{t}$, which completes the derivation.

\FloatBarrier
\bibliographystyle{IEEEtran}
\bibliography{references}

\end{document}